\documentclass{article}
\usepackage{amssymb}

\usepackage[final]{corl_2025} 
\usepackage{amsmath}
\usepackage{xcolor}
\definecolor{darkgreen}{rgb}{0,0.4,0}
\definecolor{codegreen}{rgb}{0,0.6,0}
\usepackage{amsmath, amssymb}
\usepackage{algorithm}
\usepackage[noend]{algpseudocode}
\usepackage{cleveref}
\usepackage{adjustbox} 
\usepackage{pdflscape} 
\usepackage{makecell}
\usepackage{booktabs} %

\DeclareMathOperator*{\argmin}{arg\,min}
\usepackage{graphicx}

\usepackage{multirow}
\usepackage{multicol}
\usepackage{pifont}
\newcommand{\cmark}{\text{\ding{51}}}
\newcommand{\xmark}{\text{\ding{55}}}

\usepackage[normalem]{ulem}

\usepackage{cancel}
\usepackage[font=small]{caption}
\usepackage{bbm}
\newlength\myindent
\usepackage{amsthm} 

\usepackage{fancyhdr}
\usepackage[symbol]{footmisc}
\usepackage{enumitem}
\usepackage{caption} 
\usepackage{wrapfig}
\usepackage{lipsum}
\newcommand{\connectingnode}{v_{\text{connect}}}

\newcommand{\connectorskillset}{\mathcal{C}}
\newcommand{\skillset}{\mathcal{K}}
\newcommand{\skill}{K}

\newcommand{\pskill}{K_{\text{P}}}
\newcommand{\applicabilitychecker}{\phi}

\newcommand{\qobj}{q_\text{obj}}
\newcommand{\qobjg}{q_\text{obj}^g}
\newcommand{\vobj}{\dot{q}_\text{obj}}
\newcommand{\Qobj}{Q_\text{obj}}
\newcommand{\qrobot}{q_\text{r}}
\newcommand{\vrobot}{\dot{q}_\text{r}}
\newcommand{\fsim}{f_{\text{sim}}}
\newcommand{\skillplan}{\tau_{\text{skill}}}
\newcommand{\ours}{\texttt{SPIN}}

\newcommand{\node}{v}

\newcommand{\pipost}{\pi_{\text{post}}}

\newcommand{\skillrrt}{\texttt{Skill-RRT}}
\newcommand{\lazyskillrrt}{\texttt{Lazy Skill-RRT}}
\usepackage{multicol}
\usepackage{wrapfig}

\usepackage{xr}
\title{$\texttt{SPIN}$: distilling \texttt{Skill-RRT} for long-horizon prehensile and non-prehensile manipulation}

\author{
    Haewon Jung\thanks{Equal contribution.} , Donguk Lee\footnotemark[1], Haecheol Park, JunHyeop Kim, Beomjoon Kim \\
    Korea Advanced Institute of Science and Technology (KAIST) \\
    \texttt{\{zora07, du.lee, hcp5004, amos0204, beomjoon.kim\}@kaist.ac.kr}
}

\begin{document}
\maketitle



\vspace{-5mm}
\begin{abstract}
Current robots struggle with long-horizon manipulation tasks requiring sequences of prehensile and non-prehensile skills, contact-rich interactions, and long-term reasoning.
We present \ours{} (\textbf{S}kill \textbf{P}lanning to \textbf{IN}ference), a framework that distills a computationally intensive planning algorithm into a policy via imitation learning.
We propose \texttt{Skill-RRT}, an extension of RRT that incorporates skill applicability checks and intermediate object pose sampling for solving such long-horizon problems. To chain independently trained skills, we introduce \textit{connectors}, goal-conditioned policies trained to minimize object disturbance during transitions. High-quality demonstrations are generated with \texttt{Skill-RRT} and distilled through noise-based replay in order to reduce online computation time. The resulting policy, trained entirely in simulation, transfers zero-shot to the real world and achieves over 80\% success across three challenging long-horizon manipulation tasks and outperforms state-of-the-art hierarchical RL and planning methods.
\end{abstract}
\vspace{-3mm}


\vspace{-2mm}
\section{Introduction}\label{sec:Introduction}
\begin{figure}[htb]
\vspace{-5mm}
\includegraphics[width=\textwidth]{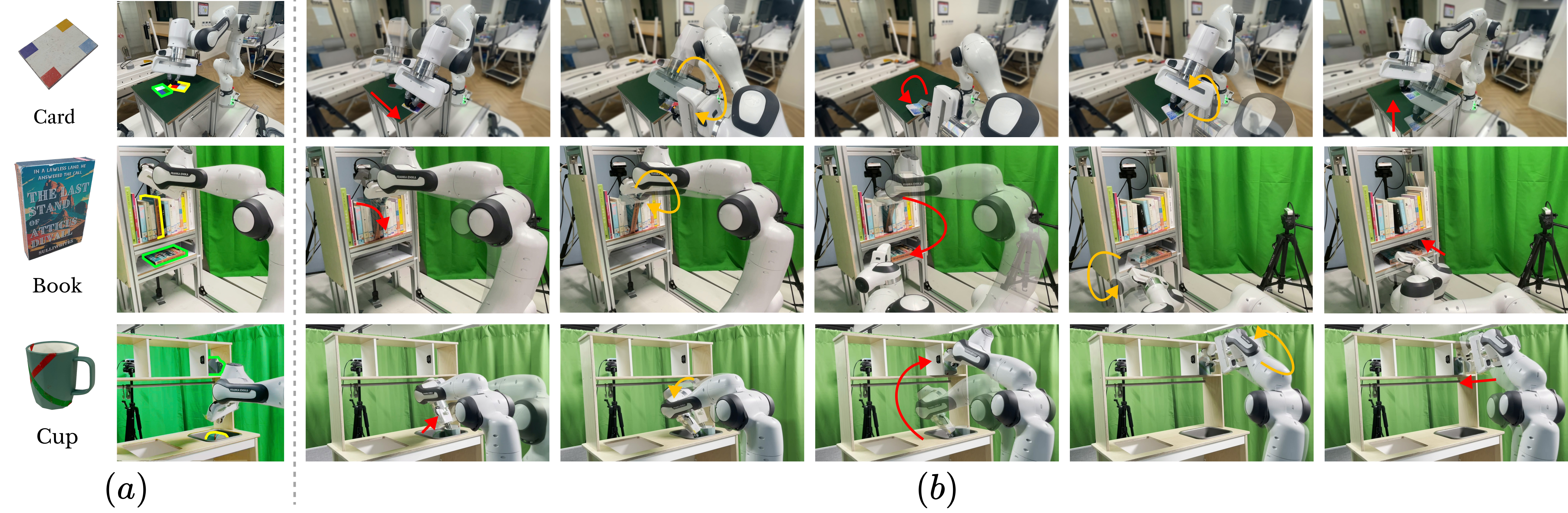}
\captionof{figure}{\label{fig:CPNP_tasks} Overview of our tasks. (a) Objects and problems. A problem is defined by initial and goal object poses marked with yellow and green. (b) Solutions for these problems. (first row) The robot must flip a thin card, initially in an ungraspable pose, by first moving it to the end of the table to make a space for grasping, flipping it, and finally sliding it to the target pose. (Second row) The robot must put the book in the lower shelf where the robot gripper cannot fit by first toppling the book to enable grasping, picking and placing it to the end of the lower shelf, and then pushing it inside. (Third row) The robot must upright the cup in a sink, grab it, place it on the cupboard, and ensure the handle is inside the desired region by re-orienting it. The red arrow indicates the movement of the object, and  the orange arrow represents the movement of the robot.}
\vspace{-8mm}
\end{figure}

\vspace{-2mm}
Humans have a remarkable capability to use sequences of prehensile (P) and non-prehensile (NP) manipulation in everyday life. For example, to take a book from a crowded shelf and place it in a tight space where our entire hand would not fit, we topple the book to enable grasping, pick-and-place it to the edge of the target space, and then push it into place using just our fingers. Our ultimate goal is to develop a policy that efficiently solves such \textit{\textbf{p}rehensile–\textbf{n}on‑\textbf{p}rehensile (PNP)} tasks that require moving an object from a random initial state to a random target pose using a sequence of P and NP manipulation skills, as illustrated in Figure~\ref{fig:CPNP_tasks}. Such problems are extremely complex: they demand reasoning not just about the feasibility of each skill but also how to \textit{chain} them, such as toppling a book to enable a grasp.
	
One way to address this problem is learning‑based approaches, such as reinforcement learning (RL) and imitation learning (IL), which have shown promise in recent years~\cite{kalashnikov2018scalable, handa2023dextreme, cho2024corn, chen2023visual, chi2023diffusion, fu2024mobile, Reuss-RSS-23, wang2023mimicplay, Zhao-RSS-23}. However, PNP problems require reasoning over a large number of actions (e.g., more than 500 joint torques) and RL struggles with such long‑horizon tasks~\cite{andrychowicz2021matters}. These issues are especially pronounced in our tasks, where the object must temporally move away from the goal before ultimately reaching it, such as sliding a card to the edge of a table to enable grasping (Figure~\ref{fig:CPNP_tasks} (b), top row). In these problems, reward design is especially difficult. IL, in contrast, sidesteps the exploration problem through dense supervision from human demonstrations. However, it scales poorly to a wide variety of initial states and goals, because collecting large‑scale, long‑horizon demonstrations is expensive~\cite{fu2024mobile, chi2024universal, lin2024learning}. 

Rather than operating in a low-level action space such as joint torques, we propose to use temporally-abstracted skills to solve PNP problems. Specifically, we leverage the recent P and NP skill learning algorithms to acquire the skills~\cite{cho2024corn, mahler2017dex, joshi2020robotic, sundermeyer2021contact, fang2023anygrasp, zhou2023learning, kim2023pre, zhou2023hacman}, and chain them together to solve PNP problems. We assume each skill has a parameterized goal-conditioned policy that takes as an input a desired object pose and state, and outputs a sequence of joint torques to achieve that pose. One way to solve PNP problems is to formulate the problem as a parameterized‑action Markov decision process (PAMDPs)~\cite{hausknecht2015deep, xiong2018parametrized, li2021hyar, dalal2021accelerating, nasiriany2022augmenting} where each parameterized action is a skill that takes in parameters, and use RL algorithms. While this shortens the planning horizon, the exploration is still challenging because PNP problems usually involve a narrow space of subgoals the policy must go through in order to achieve the goal, commonly known as \emph{narrow passages}~\cite{hsu1998finding}. Alternatively, we can use planning algorithms~\cite{lavalle2006planning}, such as RRT~\cite{lavalle1998rapidly}, which is a time-tested method for finding a very long sequence of robot configurations in a high dimensional space. However, while RRT can solve problems involving narrow passages, it typically requires a long online computation time, making it impractical in the real world.

Therefore, we propose a two-stage framework for PNP problems, called \ours{} (\textbf{S}kill \textbf{P}lanning to \textbf{IN}ference), that consists of (1) a \textit{data-generation stage} that generates a large number of solutions to PNP problems using a planner inside a simulator, and (2) a \emph{distillation stage} for learning a policy using IL to reduce online computation time and enable zero-shot real-world deployment. To do this, we propose a novel planner, \texttt{Skill-RRT}, that extends RRT~\cite{lavalle1998rapidly} into the space of skills and object goals. One challenge in designing \skillrrt{} is that typical P and NP learning algorithms are not inherently designed to learn skills in a way they would chain and there is a large gap between the resulting state of one skill and the applicable states of the next state~\cite{konidaris2009skill}. For example, in the card domain in Figure~\ref{fig:CPNP_tasks} (Top row), an NP sliding skill ends with the robot's gripper closed and on top of the card, as shown in Figure~\ref{fig:CPNP_tasks} (b) first column. However, a P skill requires the robot to move to a pre-grasp configuration at the side of the card with the gripper opened as shown in Figure~\ref{fig:CPNP_tasks} (b) second column.  One na\"ive way to solve this state gap problem is to use a collision-free motion planner to move the gripper. However, in these problems, even a small control error can disturb the object and cause it to fall, and even a slight object movement can invalidate the planned path. Instead, we need a method that intelligently breaks and makes contacts.




To address this, we introduce \textit{connectors}—goal-conditioned policies that, given the current state and goal end-effector pose, break contact and move to the goal pose with minimal disturbance to the object’s pose. For each skill, we associate a connector that brings the robot to one of the skill’s applicable states, similar to Mason's concept of “funnels”~\cite{mason1985mechanics}. We use RL to train the connectors, and to generate relevant problems for training, we propose \texttt{Lazy Skill-RRT}, which tentatively assumes the existence of connectors and teleports the robot to the next skill's applicable state. We apply \texttt{Lazy Skill-RRT} to solve PNP problems while logging the states where the robot was teleported -- indicating the need for a connector. We then train the connector on these logged problems, allowing us to focus training only on states relevant to the current problem. Equipped with connectors, we use \skillrrt{} to generate complete solutions to PNP problems.

To distill \skillrrt{} to a policy, we use Diffusion Policy \cite{chi2023diffusion} in order to cope with multi-modality in solutions. One challenge in distilling a planner to a policy, as already observed in the previous paper \cite{dalal2023imitating}, is that not all planning solutions are useful, as low-quality trajectories may degrade policy performance by leading the robot to a high-risk state. To mitigate this, we propose filtering data by replaying the \texttt{Skill-RRT} plans with noise in simulation, and discard those whose success rate is lower than a threshold. We evaluate our distilled policy both in simulation and the real world, and show that it outperforms pure planner \texttt{Skill-RRT}, a state-of-the-art PAMDPs-based RL algorithm \texttt{MAPLE}~\cite{nasiriany2022augmenting}, state-of-the-art goal-conditioned HRL algorithm \texttt{HLPS}~\cite{wang2024probabilistic}, and PPO~\cite{schulman2017proximal} in simulation, and achieves 80\% success rates in the real world in domains shown in Figure 1. Our supplementary video\footnote{Supplementary video: \texttt{cpnp.mp4}.} further illustrates the effectiveness of the proposed method in challenging environments.

\vspace{-2mm}
\section{Related Works}\label{sec:Realted_works}
\vspace{-2mm}
\paragraph{Learning prehensile and non-prehensile skills}
There are two common approaches to learning P skills. The first approach is integrating grasp pose prediction and motion planning~\cite{mahler2017dex, sundermeyer2021contact, fang2023anygrasp} where the grasp predictor outputs a grasp pose and low-level robot motion planning is used. The second approach is RL-based end-to-end policy training~\cite{kalashnikov2018scalable, joshi2020robotic, wang2022goal}. In this work, we train prehensile skills with predefined grasps using RL due to its robustness. For NP skills, recent studies use learning-based approaches where several works focus on planar pushing~\cite{ferrandis2023nonprehensile, yuan2018rearrangement, yuan2019end, peng2018sim}, some on training a policy to reposition the object to enable grasping~\cite{zhou2023learning} and some on moving the object to a target 6D pose~\cite{zhou2023hacman}. However, most of these approaches have a restricted motion repertoire, relying primarily on planar pushing or hand-crafted primitives. In this work, we use the method by \citet{kim2023pre}, an RL-based approach based on a general action space (6D end-effector pose) that automatically discovers various skills such as toppling, sliding, pushing, pulling, etc.
\paragraph{Hierarchical reinforcement learning}
A widely studied approach for HRL is the \textit{options framework}~\cite{sutton1999between,bacon2017option,harutyunyan2019termination}, where the agent learns options, each of which consists of a policy that operates over multiple steps, termination probability, and initiation set to achieve temporal abstraction. In practice, however, option learning often collapses into discovering primitive, one-step actions, failing to realize the intended temporal abstraction~\cite{singh2023pear, pateria2021hierarchical, hutsebaut2022hierarchical}. An alternative direction is \textit{goal-conditioned HRL}~\cite{wang2024probabilistic, nachum2018data,zhang2022adjacency}, where a high-level policy proposes subgoals and a low-level policy achieves them using primitive actions. While these methods have shown some promising results, they often suffer from non-stationarity during training~\cite{wang2024probabilistic}, as the high-level policy must continuously adapt to the evolving behavior of the low-level policy. These methods also face structural difficulties: the subgoals proposed by the high-level policy may be unachievable by the low-level policy. Unlike these works, we pre-train P and NP skills using existing approaches~\cite{kim2023pre, franceschetti2021robotic, makoviychuk2021isaac} to achieve temporal abstraction, rather than simultaneously learning the skills and learn to do the task.

\paragraph{Reinforcement learning with PAMDPs}
PAMDPs~\cite{hausknecht2015deep} model decision-making with a discrete set of parameterized high-level actions, each associated with its own continuous parameters (e.g., a target pose). This formulation has become increasingly viable due to recent advances in skill learning, which provide reusable low-level behaviors that generalize across a range of tasks~\cite{hausknecht2015deep, xiong2018parametrized,li2021hyar,dalal2021accelerating,nasiriany2022augmenting,wang2023task,fu2019deep,wei2018hierarchical}. However, most of these methods assume that there exist states where the terminal state of one skill can directly serve as an applicable state for the next skill. To address state gaps between skills, RAPS~\cite{dalal2021accelerating}, MAPLE~\cite{nasiriany2022augmenting}, and TRAPs~\cite{wang2023task} introduce a low-level primitive that directly controls single-timestep end-effector motions in addition to a library of skills.  However, as noted in~\cite{nasiriany2022augmenting}, training such fine-grained motion policies is challenging especially when we have a large state gap and requires a long sequence of low-level actions. In contrast, we use \lazyskillrrt{} to explicit gather problems in which state gap is an issue, and train connectors with dense reward function to address this.

\paragraph{Task and motion planning}
An alternative to learning-based skill sequencing is task and motion planning (TAMP). TAMP integrates discrete task planning with continuous motion planning to solve long-horizon problems by leveraging predefined skills with known preconditions and effects~\cite{garrett2021integrated,garrett2018ffrob, garrett2020pddlstream, kaelbling2011hierarchical, ren2024extended, migimatsu2020object, toussaint2015logic, toussaint2018differentiable}. While TAMP has demonstrated strong performance on long-horizon problems~\cite{du2023video, kim2022representation, mendez2023embodied, vu2024coast, zhu2021hierarchical}, they require symbolic representations of each skill's preconditions and effects, which are often unavailable or difficult to define in PNP settings. In particular, the NP skills typically involves multiple contact transitions and often yield stochastic outcomes where the object only approximately reaches the target pose, rendering defining effects difficult. Similar in spirit to our \skillrrt, \citet{barry2013hierarchical} propose an RRT-based skill planner that computes a collision-free object trajectory and identifies a sequence of skills to realize it. However, our method differs in three key ways: we use RL-based skills, we introduce connectors to bridge state gaps caused by the trained skills, and we reduce online computational time by distilling \texttt{Skill-RRT} into a reactive policy via IL.

\paragraph{Learning from planning solutions for manipulation}
Several works have proposed to generate IL data from planners. \citet{driess2021learning} generate TAMP datasets by using Logic-Geometric Programming (LGP), \citet{mcdonald2022guided} uses FastForward for task planning and refine motion trajectories in simulators to train hierarchical policies that imitate both task and motion planning. OPTIMUS~\cite{dalal2023imitating} uses PDDLStream~\cite{garrett2020pddlstream} to train low-level policies. We take a similar approach but use \texttt{Skill-RRT}, a planner specifically designed for PNP problems.  One core problem with planning data is that low-quality solutions can lead to failure during policy deployment~\cite{mandlekar2022matters}. OPTIMUS addresses this by filtering out trajectories where the end-effector exits a predefined workspace. While effective in simple pick-and-place domains, it is difficult to apply in contact-rich PNP problems. Instead, we evaluate the robustness of each plan by replaying it under stochastic disturbances in a simulator, keeping only plans that succeed under noise. Many works use a variety of different generative models to learn from multimodal planning solutions~\cite{chi2023diffusion,Zhao-RSS-23,dalal2023imitating, lee2024behavior,shafiullah2022behavior}, such as Conditional Variational Auto Encoders and Gaussian Mixture Models. We use Diffusion Policy~\cite{chi2023diffusion} which have proven its effectiveness in wide range of robotics applications.


\vspace{-2mm}
\section{Skill-Planning to Inference (\ours)}\label{sec:Methodology}
\vspace{-2mm}
\begin{figure*}[t] 
\centering
\vspace{-5mm}
\resizebox{\textwidth}{!}{
    \includegraphics{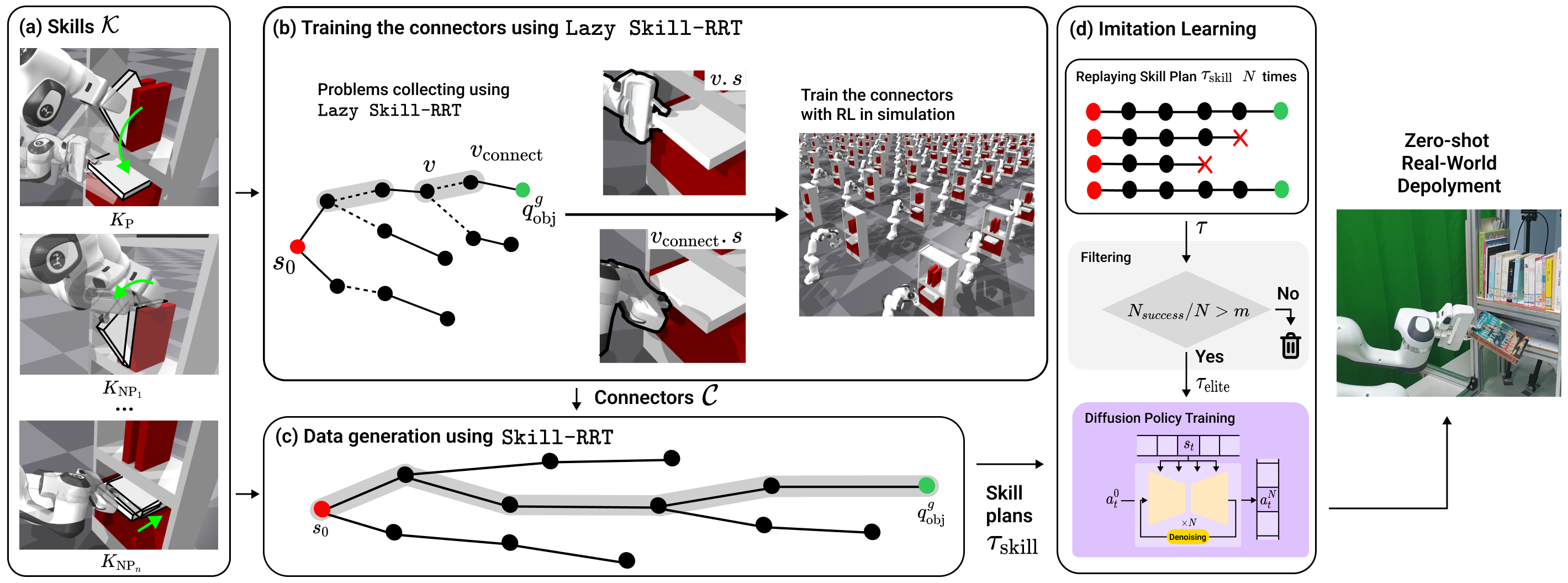}
}
\caption{Overview of \texttt{SPIN}: (a) Examples of pre-trained PNP skills in the bookshelf domain~(Figure~\ref{fig:CPNP_tasks}, row 2). Includes pick-and-place, toppling, and pushing. (b) We first use \texttt{Lazy Skill-RRT} to collect problems for training connectors. Left of (b) shows the RRT tree, where the initial and goal states are marked with red and green respectively. Each edge is defined by a skill execution, and the dotted edge and its starting and end vertices, denoted $v$ and $v_{\text{connect}}$ respectively, defines a state gap a connector needs to fill in. The middle of (b) shows the state gap. In $v.s$, the robot has just finished a prehensile skill, with the object still in between the gripper. In $v_{\text{connect}}.s$, the robot is about to begin a pushing skill to push the book into the shelf. A connector has to fill in this state gap. We collect a set of such problems, and use RL to train connectors.  (c) \texttt{Skill-RRT} is run with the trained connectors $\mathcal{C}$ and the skill library $\mathcal{K}$ to generate a skill plan $\tau_{\text{skill}}$. A skill plan consists of a sequence of skills, connectors, and their associated desired object poses, or desired robot configuration. (d) We use IL to distill skill plans into a single policy. To filter data, we replay each skill plan $N$ times, and those with a replay success rate below a predefined threshold $m$ are filtered out. The remaining high-quality trajectories are used to train a diffusion policy, which is zero-shot deployed in the real world.}\label{fig:CPNP_overview}
\vspace{-7mm}
\end{figure*}
At a high-level, \ours{} uses \texttt{Skill-RRT} to collect skill plans, and then distill it to a policy using IL. But to find the skill plans using \skillrrt, we first need connectors that allows chaining of multiple skills, and to do that, we need \lazyskillrrt~to collect relevant connector problems. See Figure~\ref{fig:CPNP_overview} for the overview of \ours. Let us begin by formally defining the PNP problem.

\subsection{PNP manipulation problem formulation and \skillrrt}\label{method:PF}

 We denote the state space as $S$, the action space as $A$. A state $s \in S$ includes object pose and velocity, robot joint positions, and velocities, denoted $\qobj, \vobj, \qrobot$ and $\vrobot$ respectively. An action $a \in A$ consists of the target end-effector pose, the gain and damping values for a low-level differential inverse kinematics (IK) controller, and the target gripper. 

For each domain, we assume we are given a set of manipulation skills, $\skillset=\{\pskill,K_{\text{NP}_1},\cdots,K_{\text{NP}_n}\}$, such as a \textit{topple} NP skill, and \emph{pick-and-place} P skill. Each skill $K$ is a tuple of two functions, $K =( \applicabilitychecker, \pi)$, where $\pi: S \times \Qobj \rightarrow A$ is a goal-conditioned policy trained using RL  that maps a state $s$ and a desired object pose $\qobj$ to an action, which is executed by a position controller, and $\Qobj \subset SE(3)$ is the space of stable object poses. The function $\applicabilitychecker:S \times \Qobj \rightarrow \{0,1\}$ is an applicability checker that tests whether it is possible to execute the skill policy in $s$ with $\qobj \in \Qobj$ as its goal, such as an IK solver that checks the existence of a feasible grasp for both the current and target object poses in a P skill.  For more details about how the skill policies are structured and learned, and how $\phi$ are defined for each skill, see Appendix~\ref{Appendix:NP_Skill}, and ~\ref{Appendix:P_Skill}.  

We assume we have a simulator $\fsim$ that takes in state $s$ and a policy conditioned on a desired object pose $\qobj$, $\pi(\cdot;\qobj)$, and simulates the policy for $N_{sim}$ number of time steps, and returns the next state, $\fsim: S \times \Pi \rightarrow S$ where $\Pi$ is the set of policies of skills defined in $\skillset$. Given a pair of an initial state $s_0 \in S$ and the \emph{ultimate} goal object pose $\qobj^g \in \Qobj$, the objective of a PNP problem is to find a sequence of skills and associated intermediate object poses, that we call a \textit{skill plan}, $\skillplan = \{\pi^{(t)},q^{(t)}_{\text{obj}}\}_{t=1}^{T}$, such that when we simulate the sequence of $T$ policies from $s_0$, the resulting state would have $\qobj^g$ as its object pose. To solve this problem, we first generate training data using \skillrrt, which we explain now.




\begin{wrapfigure}{r}{0.55\textwidth}
\vspace{-2em}
\begin{minipage}{0.55\textwidth}
\begin{algorithm}[H]
\small
\caption{\skillrrt($s_0, \qobj^g, \skillset, \connectorskillset,\Qobj$)}
\label{algo:skillrrtbackbone}
\begin{algorithmic}[1]
\State $T=\emptyset$   
\State $v_{0} \gets \{(\emptyset, \emptyset), s_0\}$ 
\State $T$.\texttt{AddNode}($parent=\emptyset, child=v_{0}$)
\For{$i = 1$ \textbf{to} $N_{\text{max}}$}
    \State ${K}, \qobj \gets $\hyperref[algo:UnifSmplSkillAndSubgoal]{\texttt{UnifSmplSkillAndSubgoal}}($\skillset,\Qobj$)
    \State ${v}_{\text{near}} \gets$ \hyperref[algo:GetFeasibleNearestNode]{\texttt{GetApplicableNearestNode}}($T, K, \qobj$)

    \If{$v_\text{near}$ is $\emptyset$} 
        \State \textbf{continue}
    \EndIf
    \State \hyperref[algo:extend]{\texttt{Extend}}$\big(T, K, v_{\text{near}}, \qobj, \connectorskillset\big)$
    \If{\texttt{NearEnough}($\qobj^g,T$)}
        \State \textbf{Return} \texttt{Retrace}($\qobj^g, T$)
    \EndIf
\EndFor
\State \textbf{Return} None
\end{algorithmic}
\end{algorithm}
\end{minipage}
\vspace{-1.5em}
\end{wrapfigure}

Algorithm~\ref{algo:skillrrtbackbone} provides the pseudocode for \skillrrt. The algorithm takes as input an initial state $s_0$, goal object configuration $\qobjg$, a skill library $\skillset$, set of connectors $\connectorskillset$, and object regions $\Qobj$. As we will soon explain, setting $\connectorskillset$ to an empty set turns the algorithm into \lazyskillrrt. The algorithm begins by initiating the tree, $T$, with an empty set (L1), defining the root node (L2), and adding it to the tree (L3). A node $v$ consists of a state $s$ and a pair of a skill policy $\pi$ and its desired object pose $\qobj$ that have been applied to the parent state to achieve the current state. For the root node, the policy and pose are set to an empty set (L2). We then begin the main for-loop. We first uniform-randomly sample a skill $K$ and the desired object pose for the skill, $\qobj$, using the function \hyperref[algo:UnifSmplSkillAndSubgoal]{\texttt{UnifSmplSkillAndSubgoal}} (L5), and compute the nearest node from the tree among the nodes where $K$ can be applied (L6). Specifically, \hyperref[algo:GetFeasibleNearestNode]{\texttt{GetApplicableNearestNode}} function returns the nearest node $v$ where $K.\phi(v.s, \qobj)$ is true, and an empty set if no such node exists. If the function returns an empty set, we discard the sample and move to the next iteration (L7-8). Otherwise, we use \hyperref[algo:extend]{\texttt{Extend}} function with $v_{near}$ (L9). Lastly, at every iteration, we check if $\qobjg$ is close enough to any of the nodes in the tree, and if it is, return the path using \texttt{Retrace} function that computes a sequence of $v$ from root node to that node (L10-11). If no such node can be found after $N_{max}$ number of iterations, we return None (L12).

Algorithm~\ref{algo:extend} shows the $\texttt{Extend}$ function. The function takes in the tree $T$, skill to be simulated $K$, node $v$ that we will use $K$ from, desired object pose we will extend to, $\qobj'$, and set of connectors $\connectorskillset$. The algorithm begins by computing a robot configuration from which $K$ is applicable, $\qrobot'$, using the function \hyperref[algo:ComputePreSkillConfig]{\texttt{ComputePreSkillConfig}} using state $v.s$ with $\qobj'$ as a goal (L1).
\begin{wrapfigure}{r}{0.5\textwidth}
\vspace{-2em}
\begin{minipage}{0.5\textwidth}
\begin{algorithm}[H]
\small
\caption{\texttt{Extend}($T, \skill, \node, \qobj', \connectorskillset$)}
\label{algo:extend}
\begin{algorithmic}[1]
\State $\qrobot' \gets \hyperref[algo:ComputePreSkillConfig]{\texttt{ComputePreSkillConfig}}(K,\node.s.\qobj,\qobj')$
\State $\connectingnode \gets \hyperref[algo:computeconnectingnode]{\texttt{ComputeConnectingNode}}(\qrobot', \skill,\connectorskillset, \node)$
\State $s' \gets \fsim(\connectingnode.s, \skill.\pi(\connectingnode.s;\qobj'))$
\If{$\hyperref[algo:Failed]{\texttt{Failed}}(s', \qobj')$}
\State \Return
\EndIf
\State $v' \gets (\skill.\pipost, \qobj',s')$
\State $T.\texttt{Add}(parent=\node, child=\connectingnode)$
\State $T.\texttt{Add}(parent=\connectingnode, child=\node')$
\end{algorithmic}
\end{algorithm}
\end{minipage}
\vspace{-1.5em}
\end{wrapfigure}
The difference between $\qrobot'$ and the robot configuration in the current state, $v.s.\qrobot$, effectively defines the state gap that our connectors must fill. To do this, we create a \emph{connecting node}, $\connectingnode$ using \hyperref[algo:computeconnectingnode]{\texttt{ComputeConnectingNode}} shown in Algorithm~\ref{algo:computeconnectingnode} (L2). When connectors $\connectorskillset$ is non-empty, this function simulates a connector and returns $\connectingnode$ whose state is the result of simulating the connector. Otherwise, it returns a node whose state is the same as the current node's state $v.s$, but whose robot configuration is set to $\qrobot'$, to effectively define the problems for training the connectors. From the connecting node's state, $\connectingnode.s$, we simulate the policy $K.\pi$ with $\qobj'$ as its goal (L3). If it fails to achieve $\qobj'$, then we return without modifying the tree (L4-5). Otherwise, we create a new node $\node'$ with the resulting state $s'$, simulated policy $\skill.\pi$, and the desired object pose $\qobj'$ (L6). We add $\connectingnode$ with its parent as $\node$, and $\node'$ with $\connectingnode$ as its parent (L7-8). For all subrotuines that have been undefined, such as \hyperref[algo:ComputePreSkillConfig]{\texttt{ComputePreSkillConfig}}, we provide their detailed pseudocodes and GPU-parallelized version used in our experiments in the Appendix~\ref{Appendix:Skill-RRT Details}.

\subsection{Training the connectors}\label{method:train_connector}

To train a set of connectors $\connectorskillset$, we collect problems using \lazyskillrrt{} by first creating a PNP problem that consists of a pair $(s_0,\qobjg)$, and then running Algorithm~\ref{algo:skillrrtbackbone} with $\connectorskillset=\emptyset$ in simulation. This will return a solution node sequence, some of which will be $\connectingnode$. We then collect a set of $(\node.s,\connectingnode.s,K)$ triplet from the node sequence, where $\node$ is the parent of $\connectingnode$, and $K$ is the skill that was used from $\connectingnode$ to get its child node for which we will train a connector. We train a connector $\pi_C$ whose goal is to go from $\node.s.\qrobot$ to the connecting node's robot configuration $\connectingnode.s.\qrobot$, using PPO~\cite{schulman2017proximal}, with minimum disturbance to the object pose in $\node.s$, for each $K$. The reward consists of four main components, $r_{\text{connector}} = r_{\text{ee}} + r_{\text{tip}} + r_{\text{obj-move}} + r_{\text{success}}$,
where $r_{\text{ee}}$ and $r_{\text{tip}}$ are dense rewards based on the distances between the current and target end-effector poses and gripper tip positions, respectively. $r_{\text{obj-move}}$ penalizes an action if it changes the object pose from previous object pose, and $r_{\text{success}}$ gives a large success reward for achieving the target end-effector pose and gripper width. More details about the implementation of the MDP formulation,
and the hyperparameters for training the connectors, can be found in Appendix~\ref{Appendix:Connector}. Having trained the connectors, now we can generate full solutions to PNP problems, which we use to train a policy.

\subsection{Distilling \skillrrt~to a policy}\label{method:IL}
While \skillrrt{} can solve PNP problems, they are computationally expensive and must compute solutions from scratch for every initial and goal states. Therefore, \ours{} distills the solutions given by \skillrrt{} to train a policy using IL, so that it generalizes to different initial and goal states without an expensive tree search. To generate data, we solve a set of PNP problems using \skillrrt{}, and then collect a data of the form $(s,a)$, where $a$ is given by each skill policy used in the skill plan, and $s$ is the state encountered during the execution of a skill plan, and we use it to train a policy. To facilitate zero-shot real-world deployment, we use a slightly different state representation from the ones defined previously. See Appendix~\ref{Appendix:imitation_learning} for details.

Distilling a planner to a policy requires solving two challenges: (1) selecting high-quality skill plans for distillation, and (2) choosing the appropriate learning algorithm that accounts for the characteristics of our data. For (1), for each skill plan, we replay it $N$ times with Gaussian or uniform random noise injected into states and joint torques, and randomize physical parameters such as friction and mass. Plans with a success rate above a threshold $m$, i.e., $\frac{N_{success}}{N} > m$, are retained. This process ensures that the dataset remains robust to perception and modeling errors and focus on solutions likely to reach the goal despite these disturbances and uncertainties. For details about the noise we used, see Appendix~\ref{Appendix:DR}. For (2), even for the same problem (i.e., the same \( s_0 \) and \( q^g_\text{obj} \)), skill plans will have different intermediate subgoals resulting in multi-modal solutions. To handle such multi-modality, we use Diffusion Policy~\cite{chi2023diffusion} as our policy. Details about the policy architecture and training hyperparameters can be found in Appendix~\ref{Appendix:imitation_learning}.


\vspace{-2mm}
\section{Experiments}\label{sec:Experiments}
\vspace{-2mm}
\subsection{Simulation results}\label{sec:main_exp}
We evaluate our method in three domains: \textit{Card Flip}, \textit{Bookshelf}, and \textit{Kitchen} (Figure~\ref{fig:CPNP_tasks}). Each domain is equipped with a set of skills $\mathcal{K}$. In Card Flip, we have $ \mathcal{K}=\{K_{\text{slide}}, K_{\text{prehensile}}\}$, where $K_{\text{slide}}$ slides the card on the table. In Bookshelf, $\mathcal{K} = \{K_{\text{topple}}, K_{\text{push}}, K_{\text{prehensile}}\}$, where $K_{\text{topple}}$ topples a book on the upper shelf, and $K_{\text{push}}$ pushes a book on the lower shelf. In Kitchen, $\mathcal{K}_{\text{kitchen}} = \{K_{\text{sink}}, K_{\text{cupboard}}, K_{\text{prehensile}}\}$, where $K_{\text{sink}}$ and $K_{\text{cupboard}}$ are non-prehensile skills that manipulates objects to the target pose in the sink and the cupboard respectively. In all domains, $K_{\text{prehensile}}$ is a pick-and-place skill that moves an object from its current pose to the target pose. We choose initial and goal object poses such that no single skill suffices, requiring combination of multiple skills to reach the goal. See Appendix~\ref{Appendix:Problem} for details of our problem distribution.

We compare our method with the following baselines:
\begin{itemize}[leftmargin=10pt, itemsep=0.3pt]
    \vspace{-2mm}
    \item PPO~\cite{schulman2017proximal}: An RL method that directly outputs joint torque commands along with joint gain and damping values without utilizing pre-trained skills $\mathcal{K}$. Trained with a reward function based on the distance between the current and goal object poses.
    \item \texttt{HLPS}~\cite{wang2024probabilistic}: A state-of-the-art hierarchical RL method that, instead of using $\mathcal{K}$, simultaneously learns low-level and high-level policies to achieve the goal. The high-level policy outputs a subgoal for a low-level policy, and is trained with the reward function based on the distance between the current and goal object pose. The low-level policy is trained with the reward function based on the distance to the subgoal provided by the high-level policy.
    \item \texttt{MAPLE}~\cite{nasiriany2022augmenting}: A state-of-the-art PAMDP RL method trained with a reward function that rewards the robot when the high-level policy selects a feasible skill, successfully completes the task, or reduces the distance between the current and goal object poses. Uses pre-trained skills $\mathcal{K}$.
    \item \texttt{Skill-RRT}: Our planner without policy distillation.
    \item \ours: Our method with the replay success threshold of $m=0.9$
    \item \ours-w/o-\texttt{filtering}: Our method without data filtering
    \vspace{-2mm}
\end{itemize}
The summary of these baselines is presented in the left of Table~\ref{table:main_exp}. The details for how we train these baselines and their MDP definitions are included in Appendix~\ref{Appendix:baseline_PPO}, ~\ref{Appendix:baseline_MAPLE}, and ~\ref{Appendix:baseline_HLPS}. For each baseline, we report the number of state-action pairs used for training across the three domains in Appendix Table~\ref{table:data_summary}. We use two metrics to evaluate the baselines: success rate (\# success episodes / \# problems attempted) and computation time. For \skillrrt, this is the average amount of plan computation time, and for all other methods, this is the sum of all inference times during the episode, averaged over all successful episodes. We use 100 problems to evaluate baselines.

Table~\ref{table:main_exp} (right) shows the results. PPO, which does not use temporally-abstracted actions, achieves zero success rate across all domains, because the tasks are too long-horizon (up to 564 number of low-level actions) for algorithms with a flat action space. \texttt{HLPS}, which simultaneously learns to do the task and discover low-level policies, also achieves zero success in all domains because of the fundamental problem with hierarchical RL: the high-level policy must adapt to the evolving low-level policy, and the training is inherently unstable. \texttt{MAPLE}, on the other hand, leverages the pre-trained skills $\mathcal{K}$, and achieves 78\% in the bookshelf domain. However, in two other domains where there is a very small region of intermediate object poses that the robot must go through to achieve the goal, such as the edge of the table for Card Flip, it achieves zero success rates (see Appendix Figure~\ref{fig:card_y_position_histograms} for a visual illustration).  In contrast, \texttt{Skill-RRT} achieves 39\%, 66\%, and 64\% success rates in the three domains. This is because it (1) leverages pre-trained skills, $\mathcal{K}$, and (2) uses sophisticated exploration strategy based on Voronoi Bias~\cite{lindemann2004incrementally} to find a path through a narrow passage. However, because of narrow passages, its average plan computation time is 85.3s, 69.2s, and 121s for the three domains, which is impractical for a real-world deployment. 

\begin{table*}[t]
\vspace{-5mm}

\begin{adjustbox}{width=\columnwidth} 

\centering
\begin{tabular}{cccc|ccccccc}
    \toprule
    & \multicolumn{3}{c}{Components} & \multicolumn{6}{c}{Problem Domain} \\
    \cmidrule(lr){2-4} \cmidrule(lr){5-10}
    & \multirow{3}{*}{Method}
    & \multirow{3}{*}{\makecell{Action \\ Type}} &
     \multirow{3}{*}{Use $\mathcal{K}$}   & 

    \multicolumn{2}{c}{Card Flip} & \multicolumn{2}{c}{Bookshelf} & \multicolumn{2}{c}{Kitchen}  \\
    & & & &
    \makecell{Success \\ rate (\%)} & \makecell{Computation \\ time (s)} &\makecell{Success \\ rate (\%)} & \makecell{Computation \\ time (s)} &\makecell{Success \\ rate (\%)} & \makecell{Computation \\ time (s)}  \\
    \cmidrule(lr){1-1} \cmidrule(lr){2-4} \cmidrule(lr){5-6}\cmidrule(lr){7-8}\cmidrule(lr){9-10}
    PPO \cite{schulman2017proximal} & 
    \textcolor{orange}{Flat RL} & 
    \textcolor{blue}{Low-level action} & 
    $\textcolor{red}{\xmark}$ &
    $0.0$ & N/A
    &
    $0.0$ &
    N/A &
    $0.0$ &
    N/A
    \\
    \midrule
    \texttt{HLPS}~\cite{wang2024probabilistic} & 
    \textcolor{red}{Hierarchical RL} & 
    \textcolor{blue}{Low-level action} & 
    $\textcolor{red}{\xmark}$ &
    $0.0$ &
    N/A &
    $0.0$ &
    N/A &
    $0.0$ & 
    N/A
    \\
    \midrule
    \texttt{MAPLE} \cite{nasiriany2022augmenting} & 
    \textcolor{red}{Hierarchical RL} & 
    \textcolor{red}{Skill $\&$ Parameter} & 
    $\textcolor{codegreen}{\cmark}$ &
    $0.0$ &
    N/A &
    $78.0$ &
    $5.3$&
    $0.0$ & 
    N/A
    \\
    \midrule

    \texttt{Skill-RRT} & 
    \textcolor{darkgreen}{Planning} & 
    \textcolor{red}{Skill $\&$ Parameter} & 
    $\textcolor{codegreen}{\cmark}$ &
    $39.0$ &
    $85.3\pm48.7$ &
    $66.0$ &
    $79.2\pm67.1$ &
    $64.0$ &
    $121\pm39.5$
    \\
    \midrule
    \texttt{SPIN}-w/o-\texttt{filtering} & 
    \textcolor{blue}{Planner distilled via IL} & 
    \textcolor{blue}{Low-level action} & 
    $\textcolor{codegreen}{\cmark}$ &
    $82.0$ & 
    $2.68\pm0.61$ &
    $83.0$ &
    $2.93\pm2.05$ &
    $87.0$ &
    $3.02\pm0.65$
    \\
    \midrule

    \makecell{\texttt{SPIN}} & 
    \textcolor{blue}{Planner distilled via IL} & 
    \textcolor{blue}{Low-level action} & 
    $\textcolor{codegreen}{\cmark}$ &
    $\textbf{95.0}$ & 
    $2.68\pm0.61$ &
    $\textbf{93.0}$ &
    $2.93\pm2.05$ &
    $\textbf{98.0}$ &
    $3.02\pm0.65$
    \\
    \bottomrule
\end{tabular}
\end{adjustbox}

\caption{Description of baselines (Left) and performance metrics (Right) with three different random seeds for each domain (standard deviations are reported in Table~\ref{table:main_exp_detail}). }
\vspace{-7mm}
\label{table:main_exp}
\end{table*}

In contrast, both variants of \ours{} achieve the highest success rates and speed among all baselines. This is because the distilled policy is trained on a diverse set of successful skill plans generated by \skillrrt, which allows it to generalize across states and recover from mistakes, leading to improved performance compared to \skillrrt \ that executes its plan directly, without a feedback loop. We also see the impact of our data filtering scheme: \ours{} has about 10 percent improvement in success rates compared to \ours-w/o-\texttt{filtering}. This is because our noise-injection-based filtering scheme eliminates skill plans that are sensitive to small deviations, such as placing the card very near the edge of the table in which even a slight error during execution would lead to failure. By filtering out such data, the policy is distilled from more robust and reproducible skill sequences, resulting in improved execution stability.

We conduct additional ablation studies on three aspects: (1) the necessity of learning the connectors, where we compare our learned connectors against the motion planner-based baseline (see supplementary video\footnote{Supplementary video: \texttt{Connector\_MP\_RL\_comparison.mp4}.}); (2) the importance of the data filtering method during imitation learning, supported by qualitative analysis; and (3) the impact of different model architectures on imitation learning performance (e.g., GMM, VAE). Detailed results and analyses are provided in Appendix~\ref{Appendix:Ablation}.

\subsection{Real World Experiments}\label{sec:real_exp}
We evaluate our distilled policy in the real world by zero-shot transferring the policy trained in simulation. For each domain, we use a set of 20 test problems consisting of different initial and goal states, and the robot must solve them in real time. The shapes of the real environments and objects are identical to those in simulation. Our policy achieves 85\% (17/20), 90\% (18/20) and 80\% (16/20) in the card flip, bookshelf, and kitchen domains respectively. There are nine failures in total, and their causes can be categorized into three cases. The first is unexpected contacts between the robot and the object during NP skill execution, which drive the object into out‑of‑distribution states (55.5\%, 5/9). In the Kitchen domain, for example, an unintended collision between the cup and the robot causes the cup to become trapped against the sink wall, preventing the robot to make further progress. The second is the hardware torque limit violations, caused by a strong collision with the environment during NP skill execution (22.2\%, 2/9). The third is geometric collisions between the object and the environment (22.2\%, 2/9), such as book colliding with the lower shelf while the robot tries to pick-and-place it to the lower shelf in the Bookshelf domain.

\vspace{-2mm}
\section{Conclusion}\label{sec:conclusion}
\vspace{-2mm}
In this work, we present \ours, consisting of: (1) \skillrrt, a planner that extends RRT for PNP problems by discovering sequences of skills and subgoals albeit at high computational cost, (2) a way to close the state gap between skills using connectors, (3) and a noise-based data filtering  and diffusion-based IL algorithms for distilling  \skillrrt{} into a policy. We show its real-world experiments demonstrating the effectiveness of our framework in three challenging domains.

\vspace{-2mm}
\section{Limitations}
\vspace{-2mm}

There are several limitations to our work. First, \texttt{SPIN} successfully generalizes over initial and goal object poses, but does not generalize over object and environment shapes. We deem this for future work.

In terms of scalability and structural limitations, although our algorithm can automatically generate high-quality skill plans (manipulation demonstrations), the data collection process remains time-consuming, and the computation cost increases with the number of skills and planning regions. We anticipate that advances in simulation and computing hardware will help alleviate this issue.

A more fundamental limitation lies in our algorithm \lazyskillrrt: the connecting node $\connectingnode$ is constructed under the assumption that the object's configuration remains unchanged from the nearest node $v$. This restricts the framework to skills with minimal terminal object motion, and prevents the use of dynamic skills such as throwing or batting. In practice, this limitation is not critical for our current tasks, as most household manipulation skills involve low-velocity object motion. Furthermore, dynamic skills are not yet supported by our simulation, and current real-world robotic platforms lack the mechanical bandwidth and control precision required for accurate execution of such fast, contact-rich behaviors.

\bibliography{references}  

\newpage
\appendix
\section{Problem Details}\label{Appendix:Problem_detail}
In this section, we provide a detailed description of the problem setup used throughout our framework. Specifically, we describe (1) how problems, defined as pairs of initial states and goal object poses, are generated; (2) how regions are defined in each domain; (3) which skills are assigned to each domain; and (4) the training and structure of the non-prehensile and prehensile skill policies in the skill library $\mathcal{K}$.

\subsection{Problem Generation}
Each PNP problem is defined by a pair consisting of an initial state \(s_0\) and a goal object pose \(q^g_\text{obj}\). The specifics of the initial state and goal object pose vary based on the environment, and this appendix provides a detailed explanation of their construction. 

To simplify the problem setup, it is assumed that both the robot and the object are stationary at the initial state. As a result, the initial velocities are set to \(s_0.\dot{q}_r = 0\) and \(s_0.\dot{q}_\text{obj} = 0\). The initial joint positions of the robot \(s_0.q_r\) are sampled randomly, but this sampling is constrained to satisfy two conditions: the positions must lie within the robot's joint limits, and there must be no collisions with the surrounding environment, including obstacles such as tables, bookshelves, kitchen furniture, or objects. The sampling of the initial object pose \( s_0.q_\text{obj} \) and the goal object pose \( q^g_\text{obj} \) is performed from \textit{distinct regions} within the environment-specific sampling spaces, ensuring that the initial and goal poses are drawn from separate regions. In particular, we intentionally sample object poses for which no collision-free and graspable robot configuration exists, i.e., there does not exist any robot configuration that can reach the object without collision and enable a feasible grasp at the sampled object pose. This ensures that the task requires, at a minimum, the use of a non-prehensile skill to reposition the object before applying a prehensile skill, followed by another non-prehensile skill if necessary. The detailed definitions of the sampling spaces for the initial and goal object poses are provided below.

\textbf{Card Flip:} The initial object pose, $s_0.q_{\text{obj}}$, is sampled such that the object is fully positioned within the boundaries of the table, regardless of whether it is flipped or unflipped. Similarly, the goal object pose, $q_{\text{obj}}^g$, is sampled entirely within the table boundaries but must have the opposite orientation from the initial object pose (i.e., if $s_0.q_\text{obj}$ is unflipped, then $q^g_\text{obj}$ must be flipped, and vice versa).

\begin{itemize}[leftmargin=10pt, itemsep=0.3pt]
    \item $s_0.q_\text{obj}$: The initial object pose is sampled from 
    
    $
    \{(x, y, z, \theta_x, \theta_y, \theta_z) \mid x \in [x^{\text{table}}_{min}, x^{\text{table}}_{max}], y \in [y^{\text{table}}_{min}, y^{\text{table}}_{max}], 
    z = z^{\text{table}}, \theta_x \in \{0, \pi\}, \theta_y = 0, \theta_z \in [0, 2\pi]\}.
    $
    
    \item $q^g_\text{obj}$: The goal object pose is sampled from 
    
    $
    \{(x, y, z, \theta_x, \theta_y, \theta_z) \mid x \in [x^{\text{table}}_{min}, x^{\text{table}}_{max}], y \in [y^{\text{table}}_{min}, y^{\text{table}}_{max}], 
    z = z^{\text{table}}, \theta_x = s_0.q_\text{obj}.\theta_x + \pi, \theta_y = 0, \theta_z \in [0, 2\pi]\}.
    $
\end{itemize}

\textbf{Bookshelf:} The initial object pose, $s_0.q_\text{obj}$,  is sampled on the upper bookshelf, with the book placed upright and inserted into the shelf. The goal object pose, $q^g_\text{obj}$, is sampled on the lower bookshelf, positioned fully inside the shelf for storage.
\begin{itemize}[leftmargin=10pt, itemsep=0.3pt]
    \item $s_0.q_\text{obj}$: The initial object pose is sampled from 
    
    $
    \{(x, y, z, \theta_x, \theta_y, \theta_z) \mid x = x^{\text{upper-shelf}}, y \in [y^{\text{upper-shelf}}_{min}, y^{\text{upper-shelf}}_{max}], 
    z = z^{\text{upper-shelf}}, \theta_x = \pi/2, \theta_y = 0, \theta_z = 0\}.
    $
    
    \item $q^g_\text{obj}$: The goal object pose is sampled from 
    
    $
    \{(x, y, z, \theta_x, \theta_y, \theta_z) \mid x = x^{\text{lower-shelf}}, y \in [y^{\text{lower-shelf}}_{min}, y^{\text{lower-shelf}}_{max}], z = z^{\text{lower-shelf}}, \theta_x = 0, \theta_y = 0, \theta_z = 0\}.
    $
    
\end{itemize}

\textbf{Kitchen:} The initial object pose, $s_0.q_\text{obj}$, is sampled within the sink. The goal object pose, $q^g_\text{obj}$, is sampled on the upper shelf, which can be either the left upper shelf or the right upper shelf, selected randomly. 

\begin{itemize}[leftmargin=10pt, itemsep=0.3pt]
    \item $s_0.q_\text{obj}$:
    Unlike objects such as cards or books, due to the asymmetry geometry of the cup, we sample $s_0.q_\text{obj}$
from a precollected dataset containing initial poses in the sink. This dataset is generated by dropping the cup into the sink from a sufficient height and recording its initial pose in simulation when it's lying down.
    \item $q^g_\text{obj}$:
    The goal object pose is sampled from 
    
    $
    \{(x, y, z, \theta_x, \theta_y, \theta_z) \mid x = x^{\text{cupboard}}, y \in \{y^{\text{l-cupboard}}, y^{\text{r-cupboard}}\}, 
    z = z^{\text{cupboard}}, \theta_x = 0, \theta_y = 0, \theta_z \in [5\pi/6, 7\pi/6]\}.
    $
    
\end{itemize}

For each problem, both the initial object pose, $s_0.q_\text{obj}$, and the goal object pose, $q^g_\text{obj}$, are randomly sampled.
\label{Appendix:Problem}

\subsection{Set of Domain Regions}
One well-known problem when planning with parameterized manipulation skills is that if you sample the object pose $\qobj$ from the full $SE(3)$ space, the probability of sampling a stable pose is zero, and planning becomes intractable~\cite{garrett2021integrated}. Therefore, following prior work~\cite{garrett2021integrated}, we associate each distinct region of stable poses with a NP skill, while the region for a P skill is defined as the union of these regions, which corresponds to $\Qobj$. For example, in the bookshelf domain, the upper shelf is associated with topple or push skills, the lower shelf with the push skill, and the P skill enables moving the book between these regions.

\begin{figure*}[h]
\centering
\resizebox{\textwidth}{!}{
    \includegraphics{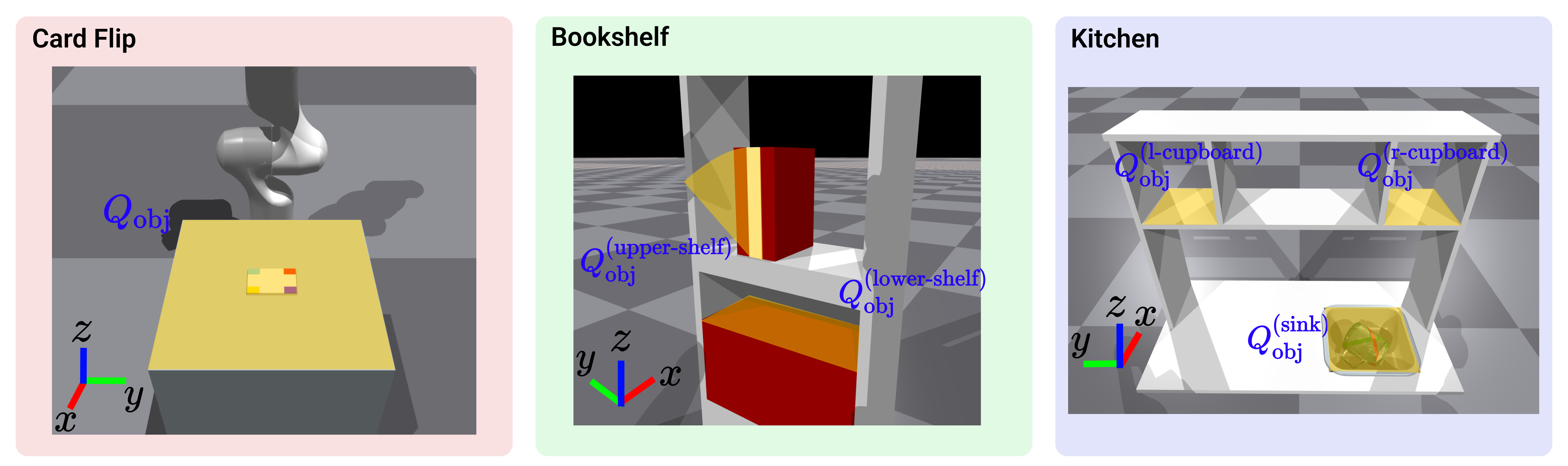}
}
\caption{Regions for each domain.}\label{fig:region}
\end{figure*}

To make planning feasible and to define the subgoal space for skill-based baselines such as \texttt{MAPLE}, we define $\Qobj$ as a subset of the $SE(3)$ space where the object can be placed stably. In the card flip domain, $\Qobj$ includes object poses lying flat on the table, regardless of whether the card is flipped. In the bookshelf domain, $\Qobj$ consists of two types: $\Qobj^\text{upper-shelf}$, where the book is fitted between other books on the upper shelf, and $\Qobj^\text{lower-shelf}$, where the book is lying on the lower shelf. In the kitchen domain, $\Qobj$ includes $\Qobj^\text{sink}$ (cup in the sink), $\Qobj^\text{l-cupboard}$ (cup on the left side of the cupboard), and $\Qobj^\text{r-cupboard}$ (cup on the right side). Figure~\ref{fig:region} illustrates examples of $\Qobj$ in each domain.

\label{Appendix:region}

\subsection{Skills For Each Domain}\label{Appendix:Skill_each_domain}
\begin{table}[ht]
\centering
\begin{adjustbox}{width=1.0\textwidth}
\begin{tabular}{|c|c|cc|cc|}
\hline
\textbf{\begin{tabular}[c]{@{}c@{}}Domain\\ names\end{tabular}} & \textbf{Card Flip}                                                                                    & \multicolumn{2}{c|}{\textbf{Bookshelf}}                                                                                                                                                                                                                                  & \multicolumn{2}{c|}{\textbf{Kitchen}}                                                                                                                                                                                                                                                                           \\ \hline
\textbf{NP skills}                                                 & $K_{\text{slide}}$                                                                                    & \multicolumn{1}{c|}{$K_{\text{topple}}$}                                                                                                      & $K_{\text{push}}$                                                                                                        & \multicolumn{1}{c|}{$K_{\text{sink}}$}                                                                                                     & $K_{\text{cupboard}}$                                                                                                                                              \\ \hline
$\phi(q,q')$                                                    & \begin{tabular}[c]{@{}c@{}}$q,q' \in \Qobj$, \\ $R_x(q) = R_x(q')$,\\ $R_y(q) = R_y(q')$\end{tabular} & \multicolumn{1}{c|}{\begin{tabular}[c]{@{}c@{}}$q,q' \in \Qobj^{\text{(upper-shelf)}}$, \\ $R_x(q) = R_x(q')$,\\ $R_z(q) = R_z(q')$\end{tabular}} & \begin{tabular}[c]{@{}c@{}}$q,q' \in \Qobj^{\text{(lower-shelf)}}$, \\ $R_x(q) = R_x(q')$,\\ $R_y(q) = R_y(q')$\end{tabular} & \multicolumn{1}{c|}{\begin{tabular}[c]{@{}c@{}}$q,q' \in \Qobj^{\text{(sink)}}$, \end{tabular}} & \begin{tabular}[c]{@{}c@{}}$q,q' \in \Qobj^{\text{(l-cupboard)}} \text{ or } \Qobj^{\text{(r-cupboard)}}$,\\ $R_x(q) = R_x(q')$,\\ $R_y(q) = R_y(q')$\end{tabular} \\ \hline
$\pi$                                                           & $\pi_{\text{slide}}$                                                                                  & \multicolumn{1}{c|}{$\pi_{\text{topple}}$}                                                                                                    & $\pi_{\text{push}}$                                                                                                      & \multicolumn{1}{c|}{$\pi_{\text{sink}}$}                                                                                                   & $\pi_{\text{cupboard}}$                                                                                                                                            \\ \hline
\end{tabular}
\end{adjustbox}
\caption{Non-prehensile manipulation skills trained using RL for each domain. The row $\phi(q,q')$ denotes the applicability checker for each skill. We write $q = \qobj$ and $\phi(q,q')$ instead of $\phi(s,q')$ with abuse of notation for brevity and clarity. Here, $R_x$, $R_y$ and $R_z$ denote the rotation matrices of pose $q$ with respect to $x,y$, and $z$ axes. For the card flip domain, slide skill is applicable if the desired pose $q'$ involves only translation and rotation wrt the z-axis from $q$. For the bookshelf domain, we have two sub-regions $\Qobj^{\text{(upper-shelf)}}$ and $\Qobj^{\text{(lower-shelf)}}$,  as shown in Figure~\ref{fig:region} (middle). To apply the topple or push skill, $q$ and $q'$ must belong to the same sub-region. Toppling skill is applicable only for orientation wrt the y-axis. The pushing skill is applicable for orientation wrt the z-axis and any translation within the lower shelf. For the kitchen domain, we have three sub-regions: $\Qobj^{\text{(sink)}}$, $\Qobj^{\text{(l-cupboard)}}$, and $\Qobj^{\text{(r-cupboard)}}$. $K_{\text{sink}}$ is applicable for any orientation or translation, as long as the object stays within the sink. $K_{\text{cupboard}}$ is applicable for an orientation wrt z-axis and any translation, as long as the object moves within a left or right cupboard. The last row, $\pi$, indicates different policies trained for different skills. The details of training are in Appendix~\ref{Appendix:NP_Skill}.}\label{tab:NP_skill_defn}
\end{table} 
\begin{table}[h]
\centering
\begin{adjustbox}{width=0.4\textwidth}
\begin{tabular}{|c|c|c|c|}
\hline
\textbf{\begin{tabular}[c]{@{}c@{}}Domain\\ names\end{tabular}}
& \textbf{Card Flip}
& {\textbf{Bookshelf}}                                                                                                        & {\textbf{Kitchen}}
\\ \hline

\textbf{P skills}
& $K_{\text{P}}$
& $K_{\text{P}}$
& $K_{\text{P}}$
\\ \hline

$\phi(q,q')$
& \multicolumn{3}{c|}{\begin{tabular}[c]{@{}c@{}}$q, q' \in \Qobj, \quad \mathcal{G}(q) \cap \mathcal{G}(q') \neq \emptyset
$\end{tabular}}

\\ \hline

$\pi$
& $\pi^\text{Card-Flip}_{\text{P}}$
& {$\pi^\text{Bookshelf}_{\text{P}}$}
& {$\pi^\text{Kitchen}_{\text{P}}$}\\ \hline
\end{tabular}
\end{adjustbox}
\caption{
Prehensile manipulation skills trained using RL for each domain. The row $\phi(q,q')$ denotes the applicability checker for each skill. We write $q = \qobj$ and $\phi(q,q')$ instead of $\phi(s,q')$ with slight abuse of notation for brevity and clarity. Specifically, $q, q' \in \Qobj$ and a skill is applicable if $\mathcal{G}(q) \cap \mathcal{G}(q') \neq \emptyset$, where $\mathcal{G}(q)$ denotes the set of feasible grasps at object pose $q$: $\mathcal{G}(q) = \{ g \in \mathcal{G}_{\text{predef}}(q) \mid \text{IK}(g) \text{ is feasible} \}.$ Here, $\mathcal{G}_{\text{predef}}(q)$ represents the set of predefined grasps relative to the object pose of $q$, and $\text{IK}(g)$ indicates whether the grasp $g$ admits a valid inverse kinematics solution. The last row, $\pi$, indicates different policies trained for different skills. Policy details and additional descriptions of $\mathcal{G}_{\text{predef}}(q)$ are provided in Appendix~\ref{Appendix:P_Skill}.
}\label{tab:P_skill_defn}
\end{table}

In this section, we describe how the skill library $\mathcal{K}$ is constructed for each domain, and introduce the structure of each skill $K$. As outlined in Table~\ref{tab:NP_skill_defn} and Table~\ref{tab:P_skill_defn}, each skill $K$ consists of two components: (1) an applicability checker $\phi$, and (2) a goal-conditioned policy $\pi$.

\underline{\textbf{Card Flip domain:}} \label{env:card_flip_skill_descript}
\begin{itemize}
        \item[] The set of skills \(\mathcal{K}\) consists of two different skills: \(\{ K_{\text{NP}_\text{slide}}, K_\text{P}\}\). \label{skill:card_flip_library}
        \item \(K_{\text{NP}_\text{slide}}\): A NP skill that manipulates a card by sliding it between two poses within the region \(Q_\text{obj}\), ensuring that the card's orientation remains unchanged along the global \( x \)- and \( y \)-axes. \label{skill:card_flip_slide}
        \begin{itemize}
            \item \( K_{\text{NP}_\text{slide}}.\phi(q_{\text{obj}}, q'_{\text{obj}}) \) determines whether an object can slide from \( q_{\text{obj}} \) to \( q'_{\text{obj}} \). It returns 1 if both poses belong to \( Q_{\text{obj}} \), and if both $\qobj$ and $q'_\text{obj}$ have identical orientations with respect to the global \( x \)- and \( y \)-axes.
        \end{itemize}
        \item \(K_\text{P}\): A P skill that moves a card within \( \Qobj \) while maintaining a grasp on it. \label{skill:card_flip_place}
        \begin{itemize}
        \item $K_\text{P}.\phi(\qobj, \qobj')$ returns 1 if \( q, q' \in \Qobj \) and there exists at least one common feasible grasp between \( q \) and \( q' \), that is, if \( \mathcal{G}(q) \cap \mathcal{G}(q') \neq \emptyset \). Otherwise, it returns 0. Here, \( \mathcal{G}(q) = \{ g \in \mathcal{G}_{\text{predef}}(q) \mid \text{IK}(g) \text{ is feasible} \} \) denotes the set of predefined grasps at pose \( q \) that admit a valid inverse kinematics solution. The predefined grasp set \( \mathcal{G}_{\text{predef}}(q) \) is defined for the card object in the card flip domain.
        \end{itemize}
\end{itemize}

\underline{\textbf{Bookshelf domain:}} \label{env:bookshelf_skill_descript}
\begin{itemize}
        \item[] The set of skills \(\mathcal{K}\) consists of three different skills: \(\{ K_{\text{NP}_\text{topple}}, K_{\text{NP}_\text{push}}, K_\text{P}\}\). \label{skill:bookshelf_library}
        \item \(K_{\text{NP}_\text{topple}}\): A NP skill that transitions a book from an upright pose to a toppled pose within the upper bookshelf region \(Q^\text{upper-shelf}_\text{obj}\), ensuring that the book maintains its alignment along the global \( x \)- and \( z \)-axes. \label{skill:bookshelf_topple}
        \begin{itemize}
            \item \( K_{\text{NP}_\text{topple}}.\phi(q_{\text{obj}}, q'_{\text{obj}}) \) determines whether an object on the upper shelf can transition between poses while toppling. It returns 1 if both poses belong to \( Q^\text{upper-shelf}_{\text{obj}} \) and if both $\qobj$ and $q'_\text{obj}$ have identical orientations with respect to the global \( x \)- and \( z \)-axes.
        \end{itemize}
        \item \(K_{\text{NP}_\text{push}}\): A NP skill that moves a book deeper into the lower bookshelf region \(Q^\text{lower-shelf}_\text{obj}\) by pushing it, ensuring that the book retains its orientation along the global \( x \)- and \( y \)-axes. \label{skill:bookshelf_push}
        \begin{itemize}
            \item \( K_{\text{NP}_\text{push}}.\phi(q_{\text{obj}}, q'_{\text{obj}}) \) determines whether an object on the lower shelf can be pushed. It returns 1 if both poses belong to \( Q^\text{lower-shelf}_{\text{obj}} \), and if both $\qobj$ and $q'_\text{obj}$ have identical orientations with respect to the global \( x \)- and \( y \)-axes.
        \end{itemize}
        \item \(K_\text{P}\): A P skill that moves a book from the upper bookshelf to the lower bookshelf while maintaining a grasp on it. \label{skill:bookshelf_place}
        \begin{itemize}
            \item \( K_{\text{P}}.\phi(\qobj, \qobj') \) is identical to \( \phi_\text{P}(\qobj, \qobj') \) in the card domain, except that the predefined grasp set \( \mathcal{G}_{\text{predef}}(q) \) is defined for the book object instead of the card.
        \end{itemize}
\end{itemize}

\underline{\textbf{Kitchen domain:}} \label{env:kitchen_skill_descript}
\begin{itemize}
        \item[] The set of skills \(\mathcal{K}\) consists of four different skills: \(\{ K_{\text{NP}_\text{sink}}, K_{\text{NP}_\text{l-cupboard}}, K_{\text{NP}_\text{r-cupboard}}, K_\text{P}\}\). \label{skill:kitchen_library}
        \item \(K_{\text{NP}_\text{sink}}\): A NP skill that manipulates a cup between two poses within the sink region \(Q_\text{obj}^\text{sink}\), ensuring that the cup remains within the boundaries of the sink throughout the movement. \label{skill:kitchen_sink}
        \begin{itemize}
            \item \( K_{\text{NP}_\text{sink}}.\phi(q_{\text{obj}}, q'_{\text{obj}}) \) determines whether an object on the sink can be manipulated. It returns 1 if both poses belong to \( Q_{\text{obj}}^{\text{sink}} \).
        \end{itemize}
        \item \(K_{\text{NP}_\text{l-cupboard}}\): A NP skill that moves a cup within the region \(Q_\text{obj}^\text{l-cupboard}\), ensuring that the cup's orientation remains unchanged along the global \( x \)- and \( y \)-axes. \label{skill:kitchen_lcupboard}
        \begin{itemize}
            \item \( K_{\text{NP}_\text{l-cupboard}}.\phi(q_{\text{obj}}, q'_{\text{obj}}) \) determines whether an object on the left cupboard can be pushed. It returns 1 if both poses belong to \( Q^\text{l-cupboard}_{\text{obj}} \), and if both $\qobj$ and $q'_\text{obj}$ have identical orientations with respect to the global \( x \)- and \( y \)-axes.
        \end{itemize}
        \item \(K_{\text{NP}_\text{r-cupboard}}\): A NP skill that moves a cup within the region \(Q_\text{obj}^\text{r-cupboard}\), ensuring that the cup's orientation remains unchanged along the global \( x \)- and \( y \)-axes. \label{skill:kitchen_rcupboard}
        \begin{itemize}
            \item \( K_{\text{NP}_\text{r-cupboard}}.\phi(q_{\text{obj}}, q'_{\text{obj}}) \) determines whether an object on the right cupboard can be pushed. It returns 1 if both poses belong to \( Q^\text{r-cupboard}_{\text{obj}} \), and if both $\qobj$ and $q'_\text{obj}$ have identical orientations with respect to the global \( x \)- and \( y \)-axes.
        \end{itemize}
        \item \(K_\text{P}\): A P skill that moves a cup from the sink and places it into a cupboard while maintaining a grasp on it. \label{skill:kitchen_place}
        \begin{itemize}
            \item \( K_\text{P}.\phi(\qobj, \qobj') \) is identical to \( \phi_\text{P}(\qobj, \qobj') \) in the card domain, except that the predefined grasp set \( \mathcal{G}_{\text{predef}}(q) \) is defined for the cup object instead of the card.
        \end{itemize}
    \end{itemize}

Next, we explain the goal-conditioned policy for each skill \( K \). Each skill has its own policy \( \pi \). The goal-conditioned policy of skill \( K \), denoted \( K.\pi \), consists of two distinct components: (1) the pre-contact policy \( K.\pi_\text{pre} \), and (2) the post-contact policy \( K.\pi_\text{post} \). 
\begin{itemize}
\item The pre-contact policy $K.\pi_\text{pre}$ computes the pre-contact robot joint positions $q'_r$ from $q_\text{obj}$ and target object poses $q'_\text{obj}$.
\item The post-contact policy $K.\pi_\text{post}$ computes a low-level action $a$, which includes changes of end-effector pose, proportional gain of joints, and damping of joints, to manipulate the object from current state $s$ and desired object pose $q'_\text{obj}$.

\end{itemize}

\subsection{Non-Prehensile Skill Policy}\label{Appendix:NP_Skill}
This section outlines the details for the non-prehensile (NP) skills policy, as described in Table~\ref{tab:NP_skill_defn}. Each skill's policy $K_\text{NP}.\pi$ consists of \( \pi_\text{pre} \) and \( \pi_\text{post} \). \( \pi_\text{post} \) computes low-level robot actions that manipulate the object toward the target object pose, using the state and target object pose as inputs. \( \pi_\text{pre} \) computes the robot configuration needed to execute \( \pi_\text{post} \), with the current and target object poses as inputs.

First we describe how we train \( \pi_\text{post} \) and then explain about \( \pi_\text{pre} \). \( \pi_\text{post} \) is trained using the Proximal Policy Optimization (PPO) algorithm, which demonstrates successful non-prehensile manipulation in \citet{kim2023pre}. To train the policy \( \pi_\text{post} \), we introduce (1) the initial problem setup and (2) the state space $S$, the action space $A$, and the reward function $R$ used for training. The problem for training post-contact policy \( \pi_\text{post} \), consist of (1) initial object pose $q^\text{init}_\text{obj}$, (2) initial robot configuration $q^\text{init}_r$, and (3) target object pose $q^\text{g}_\text{obj}$. Initial and target object poses are sampled from each region $K_\text{NP}.R$. The initial robot configuration is computed from the pre-contact policy with inputs consist of initial and target object poses $q^\text{init}_r = K.\pi_\text{pre}(q^\text{init}_\text{obj}, q^\text{g}_\text{obj})$.

For training NP skill's post-contact policy $K_\text{NP}.\pi_\text{post}$, the state space, the action space, and the reward function are defined as follows. 

\begin{itemize}[leftmargin=10pt, itemsep=0.3pt]
    \medskip
    \item \textbf{State (\( S \)):}    

\begin{table}[ht]
\centering
\begin{adjustbox}{width=0.6\textwidth} 
\begin{tabular}{|c|c|c|}
\hline
\textbf{Extract from} & \textbf{Symbol} & \textbf{Description} \\
\hline
\multirow{3}{*}{Robot configuration}  & $q^\text{(t)}_r\in\mathbb{R}^9$ &Robot joint position   \\
\cline{2-3}
 & $\dot{q}^\text{(t)}_r\in\mathbb{R}^9$ &Robot joint velocity  \\
\cline{2-3}
 & $T^\text{(t)}_\text{tool}\in\mathbb{R}^7$& Robot tool pose \\
\cline{1-3}
 Object configuration& $p^\text{(t)}_\text{obj}\in\mathbb{R}^{24}$ &Object keypoint positions \\
\hline
Action & $a^\text{(t-1)}\in\mathbb{R}^{20}$ & Previous action  \\
\hline
Target object pose & $p^\text{g}_\text{obj}\in\mathbb{R}^{24}$ &Target object keypoint positions  \\
\hline
\end{tabular}
\end{adjustbox}
\caption{The components of state space $S$ of NP skill post-contact policy $K_\text{NP}.\pi_\text{Post}$}\label{table:NP_state}
\end{table}
    The state for the NP skill's post-contact policy, \( K_\text{NP}.\pi_\text{post} \), consists of six components outlined in Table~\ref{table:NP_state}.
    The robot information includes the robot joint position $q^\text{(t)}_r$, joint velocity $\dot{q}^\text{(t)}_r$ , and tool pose $T^\text{(t)}_\text{tool}$ (The tool pose is computed from the robot joint positions $q^\text{(t)}_r$). The previous timestep action $a$ provides sufficient statistics \cite{powell2012ai}. 
    
    The object information is represented by the object keypoint positions $p^\text{(t)}_\text{obj}$ (computed from object pose $q^\text{(t)}_\text{obj}$), which represents the object's geometry. From the pre-defined relative object keypoints and the object pose $q^\text{(t)}_\text{obj}$, the object keypoint positions $p^\text{(t)}_\text{obj}$ are computed. We pre-define eight relative object keypoints (\(x, y, z\) positions) on each object's surface. For cuboid-shaped objects like a card or a book, the eight keypoints correspond to the vertices of the cuboid. For a cup, six keypoints are located on the body, and two keypoints are positioned on the handle.
    These keypoint positions are used to calculate the distances between two different object poses, both in the reward function and as state components of the skill policy. The target object pose information is represented by target object keypoint positions $p^\text{g}_\text{obj}$, which are also computed from the relative object keypoints and target object pose $q^\text{g}_\text{obj}$.

    \medskip
    \item \textbf{Action (\( A \)):} The action \( a \in A \) represents the control inputs applied to the robot.

\begin{table}[H]
\centering
\begin{adjustbox}{width=0.4\textwidth} 
\begin{tabular}{|c|c|} 
\hline
\textbf{Symbol} & \textbf{Description} \\
\hline
$\Delta q_\text{ee}\in\mathbb{R}^6$& \text{Delta end-effector pose} \\ \hline
$k_p\in\mathbb{R}^7$ & \text{Proportional gain}   \\ \hline
$\rho\in\mathbb{R}^7$ & \text{Joint damping}   \\ \hline
\end{tabular}
\end{adjustbox}
\caption{The components of action space $A$ of NP skill post-contact policy $K_\text{NP}.\pi_\text{Post}$}\label{table:NP_action}
\end{table}

    The action consists of three components. The first component, \( \Delta q_\text{ee} \), represents the delta end-effector pose, with a dimension of 6. It indicates the change in the end-effector pose, defined by the position \( (x, y, z) \) and orientation angles \( (\theta_{x}, \theta_{y}, \theta_{z}) \). The second component, \( k_p \), is the proportional gain for robot joints, which has a dimension of 7. It refers to the gain values for the robot's joints, excluding the two gripper joints. The third component, \( \rho \), represents the joint damping, and it also has a dimension of 7. It specifies the damping values for the robot's joints, excluding the gripper joints.

    For the robot joints (except gripper tip), the control process begins by computing the target end-effector pose using the end-effector pose $q^\text{(t)}_\text{ee}$ and the change of end-effector pose \( \Delta q_\text{ee} \). This target pose $q^\text{(t)}_\text{ee} + \Delta q_\text{ee}$ is then used to solve the inverse kinematics (IK) problem, yielding the target joint positions \( q_r^\text{target} \). The derivative gain \( k_d \) is calculated from the proportional gain \( k_p \) and the damping ratio \( \rho \) using the relation \( k_d = \rho \cdot \sqrt{k_p} \). Finally, the PD controller computes the torque $\tau^\text{(t)}$ at timestep $t$ as \( \tau^\text{(t)} = k_p \cdot (q_r^\text{target} - q^\text{(t)}_r) - k_d \cdot \dot{q}^\text{(t)}_r \). The computed torque $\tau$ is applied to the joints at a frequency of 100 Hz.

    For the gripper tip, no additional control is applied. The gripper width, computed by \( \pi_\text{pre} \), is maintained.

    \medskip
    \item \textbf{Reward (\( R(s_t, a_t, s_{t+1}) \)):}
    \begin{itemize}
        \item \textbf{Object Keypoint Distance Reward:} Encourages moving the object closer to its target object pose. Here, $p_\text{obj}^\text{(t)}$ 
        represents the object keypoint positions at timestep $t$ computed from $q^\text{(t)}_\text{obj}$, $p^\text{g}_\text{obj}$ 
        denotes the keypoint positions of the target object pose computed from $q^\text{g}_\text{obj}$, and $\epsilon^{\text{obj}}_0$, $\epsilon^{\text{obj}}_1$ are reward hyperparameters:
        \[
        \begin{aligned}
        r^\text{(t)}_{\text{obj}} = {\epsilon_0^{\text{obj}}\over{\|p^\text{(t)}_\text{obj} - p^\text{g}_\text{obj}\| + \epsilon^{\text{obj}}_1}}
        - {\epsilon_0^{\text{obj}}\over{\|p^\text{(t-1)}_\text{obj} - p^\text{g}_\text{obj}\| + \epsilon^{\text{obj}}_1}}
        \end{aligned}
        \]
    
        \item \textbf{Tip Contact Reward:} Encourages maintaining contact between the gripper tips and the object. Here, $p^\text{(t)}_\text{tip}$ represents the robot gripper tip positions at timestep $t$, $q^\text{(t)}_\text{obj}.\text{pos}$ represents the object position at timestep $t$, and $\epsilon^{\text{tip-obj}}_0$, $\epsilon^{\text{tip-obj}}_1$ are reward hyperparameters:
        \[
        \begin{aligned}
        r^\text{(t)}_{\text{tip-contact}} = {\epsilon_0^{\text{tip-obj}}\over{\|p_\text{tip}^\text{(t)} - q^\text{(t)}_\text{obj}.\text{pos}\| + \epsilon^{\text{tip-obj}}_1}}
        - {\epsilon_0^{\text{tip-obj}}\over{\|p^\text{(t-1)}_\text{tip} - q^\text{(t-1)}_\text{obj}.\text{pos}\| + \epsilon^{\text{tip-obj}}_1}}
        \end{aligned}
        \]
    
    \item \textbf{Success Reward:} A success reward, \( r_{\text{succ}} \), is given when the object is successfully manipulated to the target object pose \( q^\text{g}_\text{obj} \); otherwise, the reward is 0. The distance function between two object poses, \( q^1_\text{obj} \) and \( q^2_\text{obj} \), is defined as \( d_{SE(3)}(q^1_\text{obj}, q^2_\text{obj}) = \Delta T(q^1_\text{obj}, q^2_\text{obj}) + \alpha \Delta\theta(q^1_\text{obj}, q^2_\text{obj}) \), where \( \Delta T \) represents the positional difference, \( \Delta\theta \) denotes the orientational difference, and \( \alpha \) is a weighting factor for the orientation difference, set to \( 0.1 \). The object manipulation is considered successful if the distance between the current and target object poses is less than \( \delta_\text{obj} \).

    \[
    r^\text{(t)}_{\text{success}} =
    \begin{cases} 
    r_{\text{succ}} & \text{if } d_{SE(3)}(q^\text{(t)}_{\text{obj}}, q^\text{g}_{\text{obj}}) < \delta_\text{obj}, \\
    0 & \text{otherwise}.
    \end{cases}
    \]

    \end{itemize}

    The overall reward is defined as:
    \[
    r^\text{(t)}_{\text{NP}} = r^\text{(t)}_{\text{obj}} + r^\text{(t)}_{\text{tip-contact}} + r^\text{(t)}_{\text{success}}
    \]
    The hyperparameters of the reward function, $\epsilon_0^{\text{obj}}$, $\epsilon_1^{\text{obj}}$, $\epsilon_0^{\text{tip-obj}}$, $\epsilon_1^{\text{tip-obj}}$, $r_{\text{succ}}$, and $\delta_\text{obj}$, vary depending on the specific NP skill being trained. The values of these reward hyperparameters for each NP skill \( K_{\text{NP}_i} \) are provided in Table~\ref{table:NP_reward}.

\end{itemize}

\begin{table}[H]
\centering
\begin{adjustbox}{width=0.7\textwidth} 
\begin{tabular}{|l|c|c|c|c|c|c|}
\hline
\textbf{Domain} & \textbf{Card Flip} & \multicolumn{2}{c|}{\textbf{Bookshelf}} & \multicolumn{3}{c|}{\textbf{Kitchen}} \\ \hline
\textbf{Skills} & slide & topple & push & sink & l-cupboard & r-cupboard \\ \hline
$\epsilon_0^{\text{obj}}$       & 0.02 & 0.3 & 1.0  & \multicolumn{3}{c|}{0.2} \\ \hline
$\epsilon_1^{\text{obj}}$       & \multicolumn{2}{c|}{0.02} & 0.01 & \multicolumn{3}{c|}{0.02} \\ \hline
$\epsilon_0^{\text{tip-obj}}$ & 0.0 & \multicolumn{2}{c|}{1.0} & \multicolumn{3}{c|}{0.3} \\ \hline
$\epsilon_1^{\text{tip-obj}}$ & 0.0 & 0.02 & 0.01 & \multicolumn{3}{c|}{0.02} \\ \hline
$r_{\text{succ}}$ & 1000 & 500 & 50  & 500 & \multicolumn{2}{c|}{1,000} \\ \hline
$\delta_{\text{obj}}$ & \multicolumn{6}{c|}{0.005}  \\ \hline
\end{tabular}
\end{adjustbox}
\caption{Reward hyperparameter values for training NP skill policies}\label{table:NP_reward}
\end{table}

For training pre-contact policy $K_\text{NP}.\pi_\text{pre}$, the state space consist of two object poses, initial object pose $q_\text{obj}$, and target object pose $ q^\text{g}_\text{obj}$. The pre-contact policy outputs $q'_r$, which is the robot joint positions for initiating the $K_\text{NP}$. The pre-contact policy is jointly trained with $\pi_\text{post}$ following the algorithm in \citet{kim2023pre}.

NP skill policies utilize a multilayer perceptron (MLP) architecture to generate low-level robot actions based on the state and target object pose information. The post-contact policy $\pi_\text{post}$ employs a five-layer MLP. The MLP has hidden dimensions of $[512, 256, 256, 128]$, input dimension 93, and an output dimension of 20. ELU is used as the activation function for the hidden layers, while Identity is applied as the final activation function. Pre-contact policy $\pi_\text{pre}$ employs a five-layer MLP. The MLP has hidden dimensions of $[512, 256, 256, 128]$, input dimension 14, and an output dimension of 9. ELU is used as the activation function for the hidden layers, while Identity is applied as the final activation function. We summarize the hyperparamters of architecture in Table \ref{table:NP_arch}

\begin{table}[H]
    \fontsize{8}{8}\selectfont
    \centering
    
\begin{tabular}{c|c|c|c|c|c}
    \toprule
    & \makecell{input\\dimensions}& \makecell{hidden\\dimensions} & \makecell{output\\dimensions} &\makecell{hidden\\activations} &\makecell{output\\activation} \\
    \cmidrule(lr){1-1}\cmidrule(lr){2-2}\cmidrule(lr){3-3}\cmidrule(lr){4-4}\cmidrule(lr){5-5}\cmidrule(lr){6-6}
    $\pi_\text{post}$
    & 93
    & \multirow{2}{*}{$[512, 256, 256, 128]$}
    & 20 
    &
    \multirow{2}{*}{ELU} & 
    \multirow{2}{*}{Identity}
    \\
    \cmidrule(lr){1-1}\cmidrule(lr){2-2}\cmidrule(lr){4-4}
    $\pi_{\text{pre}}$
    & 14
    & 
    & 9
    &
    &
    \\
    \bottomrule
    \end{tabular}
    \caption{The network architecture of NP skill policies}
    \label{table:NP_arch}
\end{table}

\subsection{Prehensile Skill Policy}\label{Appendix:P_Skill}

This section describes the details of the prehensile (P) skill policy, following the structure in Table~\ref{tab:P_skill_defn}. Each skill's policy $K_\text{P}.\pi$ consists of two components: the pre-contact policy $\pi_\text{pre}$ and the post-contact policy $\pi_\text{post}$.

To enable $\pi_\text{pre}$ to generate a feasible robot configuration for grasping, we first construct a set of predefined grasp poses $\mathcal{G}_\text{predef}$ (mentioned in Table~\ref{tab:P_skill_defn}) for each object. These grasps provide candidate end-effector poses the robot can use to establish contact with the object before executing $\pi_\text{post}$. For cuboid objects such as the card and the book, the predefined grasp poses are generated as follows: (1) we form multiple contact points on a rectangle slightly offset from the edges on the broad surface of the object, (2) we set the grasp direction from each contact point toward the object's center of mass, and (3) we compute the grasp pose based on the contact point and the grasp direction. The grasp width is set to the thickness of the cuboid object. For the cup object, we adopt the grasp generation method from ACRONYM~\cite{eppner2021acronym} due to the cup's asymmetric geometry. Specifically, (1) we sample random grasp poses, (2) check for collisions between the cup and the gripper, (3) close and shake the gripper, and (4) if the object remains grasped after shaking, we save the grasp pose and grasp width.

The pre-contact policy $\pi_\text{pre}$ is implemented as a heuristic function that generates a robot configuration to establish a grasp on the object before executing $\pi_\text{post}$. Given the current object pose and the target object pose, $\pi_\text{pre}$ first checks the applicability condition through $\phi(q, q')$. If $\phi(q, q') = 1$, meaning there exists a common feasible grasp between $q$ and $q'$, it randomly selects a grasp from the intersection $\mathcal{G}(q) \cap \mathcal{G}(q')$ and solves for a collision-free inverse kinematics (IK) solution. If multiple grasps are available, one is selected randomly. If $\phi(q, q') = 0$, $\pi_\text{pre}$ instead randomly selects a grasp from the feasible grasp set $\mathcal{G}(q)$ associated with the initial object pose.

The post-contact policy $\pi_\text{post}$ computes low-level robot actions to manipulate the object from its current pose toward the desired pose. It takes as input the robot's state and the target object pose. The policy $\pi_\text{post}$ is trained using the Proximal Policy Optimization (PPO) algorithm. The training problem for $\pi_\text{post}$ consists of (1) an initial object pose $q^\text{init}_\text{obj}$, (2) an initial robot configuration $q^\text{init}_r$, and (3) a target object pose $q^\text{g}_\text{obj}$. The initial and goal object poses are sampled from the valid region $\Qobj$, and the initial robot configuration $q^\text{init}_r$ is computed using the pre-contact policy as $q^\text{init}_r = \pi_\text{pre}(q^\text{init}_\text{obj}, q^\text{g}_\text{obj})$.

For training the post-contact policy $\pi_\text{post}$, the state space, the action space, and the reward function are defined as follows. 

\begin{itemize}[leftmargin=10pt, itemsep=0.3pt]
    \medskip
    \item \textbf{State (\( S \)):}

\begin{table}[H]
\centering
\begin{adjustbox}{width=0.7\textwidth} 
\begin{tabular}{|c|c|c|}
\hline
\textbf{Extract from} & \textbf{Symbol} & \textbf{Description}  \\
\hline
\multirow{5}{*}{Robot configuration}  
\text{ } & $q^\text{(t)}_r\in\mathbb{R}^9$ & Robot joint position   \\
\cline{2-3}
 & $\dot{q}^\text{(t)}_r\in\mathbb{R}^9$ & Robot joint velocity  \\
\cline{2-3}
 & $T^\text{(t)}_\text{tool}\in\mathbb{R}^7$ & Robot tool pose  \\
\cline{2-3}
 & $p^\text{(t)}_\text{ee}\in\mathbb{R}^{24}$ & Robot end-effector keypoint positions  \\
\cline{2-3}
 & $p^\text{(t)}_\text{tip}\in\mathbb{R}^6$ & Robot gripper tip positions\\
 \cline{1-3}
Object configuration & $p^\text{(t)}_\text{obj}\in\mathbb{R}^{24}$ & Object keypoint positions  \\
\hline
Action & $a^\text{(t-1)}\in\mathbb{R}^{20}$ & Previous action \\
\hline
Target object pose & $p^\text{g}_\text{obj}\in\mathbb{R}^{24}$ & Target object keypoint positions  \\
\hline
\end{tabular}
\end{adjustbox}
\caption{The components of state space $S$ of P skill policy $\pi_\text{P}$}\label{table:P_state}
\end{table}

    The state space consists of eight components outlined in Table~\ref{table:P_state}. 
     The end-effector pose is computed from the robot joint positions \( q^\text{(t)}_r \). Eight keypoints are defined at the vertices of the end-effector. Using these keypoints and the end-effector pose, the end-effector keypoint positions \( p^\text{(t)}_\text{ee} \) are computed. Similarly, the gripper tip positions \( p^\text{(t)}_\text{tip} \) are computed from the robot joint positions \( q^\text{(t)}_r \). The remaining state components are identical to the non-prehensile skill's post-contact policy's state components.
    

    \medskip
    
    \item \textbf{Action (\( A \)):} The action space of the prehensile skill is identical to that of the NP skill. For the gripper tip, control is applied with a width of 0 to maintain the grasp on the object.

    \medskip
    
    \item \textbf{Reward (\( R(s_t, a_t, s_{t+1}) \)):}
    \begin{itemize}
        \item \textbf{Object Keypoint Distance Reward:} 
        The reward $r^\text{(t)}_\text{obj}$ computation is identical to the object keypoint distance reward used in non-prehensile post-contact policy training.
    
        \item \textbf{Object Rotation Reward:} Encourages aligning the object orientation with its target object orientation. Here, $q^\text{(t)}_\text{obj}.\theta$ represents the object pose orientation at timestep $t$, and $q^\text{g}_\text{obj}.\theta$ denotes the target object pose orientation:
        \[
        \begin{aligned}
        r^\text{(t)}_{\text{rot}} = {\epsilon_0^{\text{rot}}\over{\|q^\text{(t)}_\text{obj}.\theta - q^\text{g}_\text{obj}.\theta\| + \epsilon^{\text{rot}}_1}}
        - {\epsilon_0^{\text{rot}}\over{\|q^\text{(t-1)}_\text{obj}.\theta - q^\text{g}_\text{obj}.\theta\| + \epsilon^{\text{rot}}_1}}
        \end{aligned}
        \]
    
        \item \textbf{Relative Grasp Reward:} Ensures the gripper tips maintain a consistent grasp on the object. Here, \( p^\text{(t)}_\text{ee-rel} \) represents the relative position of the robot end-effector keypoints. It is computed from the absolute end-effector keypoints \( p^\text{(t)}_\text{ee} \) with respect to the object pose \( q^\text{(t)}_{\text{obj}} \) at timestep \( t \):
        \[
        \begin{aligned}
        r^\text{(t)}_{\text{grasp}} = w^{\text{grasp}}\|p^\text{(t)}_\text{ee-rel}- p^\text{(t-1)}_\text{ee-rel}\| 
        \end{aligned}
        \]
    
    \item \textbf{Success Reward:}
    The reward $r^\text{(t)}_\text{success}$ computation is identical to the success reward used in non-prehensile post-contact policy training. 

    
    \end{itemize}

    The overall reward is defined as:
    \[
    r_{\text{P}} = r^\text{(t)}_{\text{obj}} + r^\text{(t)}_{\text{rot}} + r^\text{(t)}_{\text{grasp}} + r^\text{(t)}_{\text{success}}
    \]

    The hyperparameters of the reward function, $\epsilon_0^{\text{object}}$, $\epsilon_1^{\text{object}}$, $\epsilon_0^{\text{rotation}}$, $\epsilon_1^{\text{rotation}}$, $w^{\text{grasp}}$, $r_{\text{succ}}$, and $\delta_\text{obj}$, vary depending on the specific P skill being trained. The values of these reward hyperparameters for each P skill \( K_\text{P} \) are provided in Table~\ref{table:P_reward}.

\end{itemize}

\begin{table}[H]
\centering
\begin{adjustbox}{width=0.4\textwidth} 
\begin{tabular}{|l|c|c|c|}
\hline
\textbf{Domain} & \multicolumn{1}{c|}{\textbf{Card Flip}} & \multicolumn{1}{c|}{\textbf{Bookshelf}} & \multicolumn{1}{c|}{\textbf{Kitchen}} \\ \hline
$\epsilon_0^{\text{object}}$       & 0.2 & 0.15 & 1 \\ \hline
$\epsilon_1^{\text{object}}$       & \multicolumn{2}{c|}{0.02} & 0.015 \\ \hline
$\epsilon_0^{\text{rotation}}$       & 0.0 & 0.15 & 0.0  \\ \hline
$\epsilon_1^{\text{rotation}}$       & 0.0 & 0.1 & 0.0 \\ \hline
$w^{\text{grasp}}$       & 0.0 & 10.0 & 15.0 \\ \hline
$r_{\text{succ}}$ & \multicolumn{3}{c|}{1000} \\ \hline
$\delta_{\text{obj}}$ & \multicolumn{3}{c|}{0.005}  \\ \hline
\end{tabular}
\end{adjustbox}
\caption{Reward hyperparameter values for training P skill post-contact policy}\label{table:P_reward}
\end{table}

We train the P skill policy with Proximal Policy Optimization (PPO) \cite{schulman2017proximal}.

\medskip
P skill policies utilize a multilayer perceptron (MLP) architecture to generate low-level robot actions based on the state and target object pose information. \( K_{\text{P}}.\pi_{\text{P}} \) employs a five-layer MLP with an input dimension of 123. Other components of the network architecture are identical to the NP skill post-contact policy's network architecture. The hyperparameters of network is summarized in Table \ref{table:P_arch}
\begin{table}[H]
    \fontsize{8}{8}\selectfont
    \centering
    
\begin{tabular}{c|c|c|c|c|c}
    \toprule
    & \makecell{input\\dimensions}& \makecell{hidden\\dimensions} & \makecell{output\\dimensions} &\makecell{hidden\\activations} &\makecell{output\\activation} \\
    \cmidrule(lr){1-1}\cmidrule(lr){2-2}\cmidrule(lr){3-3}\cmidrule(lr){4-4}\cmidrule(lr){5-5}\cmidrule(lr){6-6}
    $\pi_\text{post}$
    & 123
    & $[512, 256, 256, 128]$
    & 20
    & ELU
    & Identity
    \\

    \bottomrule
    \end{tabular}
    \caption{The network architecture of P skill post-contact policy}
    \label{table:P_arch}

\end{table}

\newpage
\section{Detail of \texttt{SPIN}}\label{Appendix:Detailed_algorithm}
In this section, we describe (1) additional algorithms used in \texttt{Skill-RRT}, (2) the training procedure of the connector policy, (3) the state-action structure and network architecture used for imitation learning in \texttt{SPIN}, and (4) the domain randomization ranges in simulation, along with the level of noise added to the observations for real-world deployment.

\subsection{Detail of \texttt{Skill-RRT}}\label{Appendix:Skill-RRT Details}
\newcommand{\gpuskillrrt}{\texttt{Skill-RRT-Batch}}
This section provides the algorithms used in \texttt{Skill-RRT}. These include \texttt{ComputeConnectingNode}, \texttt{UnifSmplSkillAndSubgoal}, \texttt{GetApplicableNearestNode}, and \texttt{Failed}. Then, we describe \gpuskillrrt, an extended \texttt{Skill-RRT} algorithm that extends nodes in parallel to accelerate planning and data collection times using GPU-based simulation, specifically Isaac Gym \cite{makoviychuk2021isaac}. Note that we use \gpuskillrrt~instead of the vanilla \texttt{Skill-RRT} throughout our experiments.

Across the algorithms, the distance function \( d_{SE(3)} \), which is used to determine the success of NP and P skills, is used in the same way.

Algorithm \ref{algo:UnifSmplSkillAndSubgoal} provides the pseudocode for the \texttt{UnifSmplSkillAndSubgoal} function. This function takes as input the set of skills $\mathcal{K}$, and the object regions $Q_\text{obj}$. It outputs a uniformly sampled skill $K$ and the desired object pose for the skill, $q_\text{obj}$. The algorithm begins by uniformly sampling a skill $K$ from the set of skills $\mathcal{K}$ (L1). Then, the desired object pose $q_\text{obj}$ is set to the goal object pose $q^g_\text{obj}$ with a goal sampling probability $p_g$ (L2–4), or it is uniformly sampled from the object regions $Q_\text{obj}$ otherwise (L5–6). Finally, the algorithm returns the sampled skill $K$ and the desired object pose for the skill $q_\text{obj}$.
\begin{center}
\begin{minipage}{0.6\textwidth}
\begin{algorithm}[H]
\caption{\texttt{UnifSmplSkillAndSubgoal}($\mathcal{K}, Q_{\text{obj}}$)}\label{algo:UnifSmplSkillAndSubgoal}
\begin{algorithmic}[1]

\State $K \gets \text{UniformSample}(\mathcal{K})$
\State $p \leftarrow \text{rand}(1)$ 
\If{$p < p_g$}
    \State $q_{\text{obj}} \gets q^{g}_\text{obj}$
\Else
    \State $q_{\text{obj}} \gets \text{UniformSamplePose}(Q_{\text{obj}})$
\EndIf

\State \Return $K, q_{\text{obj}}$

\end{algorithmic}
\end{algorithm}
\end{minipage}
\end{center}

Algorithm \ref{algo:GetFeasibleNearestNode} provides the pseudocode for the \texttt{GetApplicableNearestNode} function. It takes in the tree $T$, the skill $K$ for which feasibility is to be checked, and the desired object pose for the skill, $q_{\text{obj}}$. If the skill $K$ is a non-prehensile skill, we compute all feasible nodes for the given skill and desired object pose in the tree using the feasibility checker $K.\phi(s, q_{\text{obj}})$. We then find the nearest node, $v_{\text{near}}$, to the desired object pose $q_{\text{obj}}$ among the feasible nodes using the distance function $d_{SE(3)}$, and return $v_{\text{near}}$ (L1-3). If the skill $K$ is a prehensile skill, we first compute the closest node, $v_{\text{close}}$, to the desired object pose $q_{\text{obj}}$ in the tree (L5). If the closest node $v_{\text{close}}$ passes the feasibility check using the feasibility checker for the prehensile skill, $v_{\text{near}}$ is set to $v_{\text{close}}$ (L6-7). Otherwise, $v_{\text{near}}$ is set to the empty set (L8-9). The different order of finding the nearest node and performing the feasibility check for prehensile skills is due to the computational expense of the feasibility checker of the prehensile skill, as it involves collision checking.
\begin{center}
\begin{minipage}{0.6\textwidth}
\begin{algorithm}[H]
\caption{\texttt{GetApplicableNearestNode}($T, K, \qobj$)}\label{algo:GetFeasibleNearestNode}
\begin{algorithmic}[1]

\If{$K \in \{K_{\text{NP}_1}, \dots, K_{\text{NP}_n}\}$}
    \State $V_{\text{feasible}} = \{v \mid v \in T.V, K.\phi(v.s, q_{\text{obj}})==1 \}$
    \State $v_{\text{near}} \gets \argmin_{v \in V_{\text{feasible}}} d_{SE(3)}(v.q_{\text{obj}}, {q}_{\text{obj}})$
\Else \Comment{If prehensile skill}
    \State $v_{\text{close}} \gets \argmin_{v \in T.V} d_{SE(3)}(v.q_{\text{obj}}, {q}_{\text{obj}})$
    \If{$K.\phi(v_{\text{close}}.s,q_{\text{obj}})$}
        \State $v_\text{near} \gets v_{\text{close}}$
    \Else
        \State $v_{\text{near}} \gets \emptyset$
    \EndIf
\EndIf
\State \Return $v_{\text{near}}$

\end{algorithmic}
\end{algorithm}
\end{minipage}
\end{center}

Algorithm~\ref{algo:computeconnectingnode} shows the $\texttt{ComputeConnectingNode}$ function. The function takes as input the target robot joint configuration \( q_r' \), the skill \( K \) to be simulated, the connector set \( \mathcal{C} \) which includes connector policies, and the node \( v \) from which the connector policy or teleportation is applied.
The algorithm begins by checking if $\connectorskillset$ is empty, in which case this will instantiate \lazyskillrrt (L1). In this case, we create a connecting node $\connectingnode$ with empty skill and object pose, with the same state as $v$, except the robot joint configuration is teleported to $\qrobot'$ (L2-4). Otherwise, we retrieve the connector $\pi_C$ for the given skill $K$ (L6), simulate it from $v.s$ with pre-contact configuration $\qrobot'$ as a goal (L7), and create $\connectingnode$ using $\pi_C$, $\qrobot'$, and the resulting state $s'$ (L8).
\begin{center}
\begin{minipage}{0.6\textwidth}
\begin{algorithm}[H]
\small
\caption{\texttt{ComputeConnectingNode}($\qrobot', \skill, \connectorskillset, \node$)}\label{algo:computeconnectingnode}
\begin{algorithmic}[1]
\State IsLazy $\gets \connectorskillset == \emptyset$
\If{IsLazy}
\State $\connectingnode \gets (\emptyset, \emptyset, \node.s)$
\State $\connectingnode.s.\qrobot \gets \qrobot'$ \Comment{Teleport the $q_r$ to $q_r'$}
\Else
\State $\pi_C \gets \texttt{GetConnectorForSkill}(\connectorskillset,\skill) $
\State $s' \gets \fsim(\node.s, \pi_C(v.s;\qrobot'))$
\State $\connectingnode \gets (\pi_C,\qrobot', s')$
\EndIf
\Return $\connectingnode$
\end{algorithmic}
\end{algorithm}
\end{minipage}
\end{center}

Algorithm~\ref{algo:ComputePreSkillConfig} describes the \texttt{ComputePreSkillConfig} function. This function takes as input a skill \( K \), the current object pose \( \node.s.\qobj \), and the target object pose \( \qobj' \). The goal is to compute a pre-contact robot joint configuration that positions the robot appropriately to execute the skill \( K \) starting from \( \node.s.\qobj \) and targeting \( \qobj' \). The algorithm first retrieves the pre-contact policy \( K.\pi_\text{pre} \) associated with the skill \( K \) (L1). It then applies this policy to the object pose transition \((\node.s.\qobj, \qobj')\) to obtain the corresponding robot configuration \( \qrobot' \) (L2), which serves as the starting point for the skill execution. The computed configuration \( \qrobot' \) is returned as the output.
\begin{center}
\begin{minipage}{0.6\textwidth}
\begin{algorithm}[H]
\small
\caption{\texttt{ComputePreSkillConfig}($K,\node.s.\qobj,\qobj'$)}\label{algo:ComputePreSkillConfig}
\begin{algorithmic}[1]

\State $K.\pi_\text{pre} \gets \texttt{GetPreContactPolicy}(K)$
\State $\qrobot' \gets K.\pi_\text{pre}(\node.s.\qobj,\qobj')$

\Return $\qrobot'$
\end{algorithmic}
\end{algorithm}
\end{minipage}
\end{center}

Algorithm \ref{algo:Failed} checks whether a skill fails given the state $s$ and the desired object pose $q_{\text{obj}}$. If the distance between the object pose in the state $s.q_{\text{obj}}$ and the desired object pose $q_{\text{obj}}$ is less than the predefined threshold $\delta_{\text{obj}}$, the algorithm returns false. Otherwise, it returns true.
\begin{center}
\begin{minipage}{0.5\textwidth}
\begin{algorithm}[H]
\caption{\texttt{Failed}($s, q_\text{obj}$)}\label{algo:Failed}
\begin{algorithmic}[1]

\If{$d_{SE(3)}(s.q_\text{obj},q_\text{obj})<\delta_\text{obj}$}
    \State $fail \gets False$
\Else
    \State $fail \gets True$
\EndIf

\State \Return $fail$

\end{algorithmic}
\end{algorithm}
\end{minipage}
\end{center}

We now describe \gpuskillrrt~in Algorithm \ref{algo:gpuskillrrtbackbone}, a parallelized version of the original \texttt{Skill-RRT}. The inputs and outputs are the same as those of the original \texttt{Skill-RRT}. The main differences in \gpuskillrrt~are: 1) uniformly sampling skills and desired object poses in batches using the \texttt{UnifSmplSkillAndSubgoalBatch} function (L5); 2) finding the feasible and nearest node $\mathbf{v}_\text{near}$ using the \texttt{GetApplicableNearestNodeBatch} function (L6); and 3) extending nodes in parallel with GPU-based simulation via the \texttt{ExtendBatch} function. Bold letters indicate a batch of components. We do not rewrite the batched versions of the functions because single components (normal letters) are simply replaced by batches of components (bold letters).
\begin{center}
\begin{minipage}{0.7\textwidth}
\begin{algorithm}[H]
\caption{\texttt{\gpuskillrrt}($s_0, \qobj^g, \skillset, \connectorskillset,\Qobj$)}
\begin{algorithmic}[1]
\State $T=\emptyset$   
\State $v_{0} \gets \{(\emptyset, \emptyset), s_0\}$ 
\State $T$.\texttt{AddNode}($parent=\emptyset, child=v_{0}$)

\For{$i = 1$ \textbf{to} $N_{\text{max}}$}
    \State $\mathbf{K}, \mathbf{\qobj} \gets ${\texttt{UnifSmplSkillAndSubgoalBatch}}($\skillset, \Qobj$)
    \State $\mathbf{{v}_{\text{near}}} \gets$ {\texttt{GetApplicableNearestNodeBatch}}($T, \mathbf{K}, \mathbf{\qobj}$)
    \If{$\mathbf{v_\text{near}}$ is $\emptyset$} 
        \State \textbf{continue}
    \EndIf
    \State \texttt{ExtendBatch}$\big(T, \mathbf{v_{\text{near}}}, \mathbf{K}, \mathbf{\qobj}, \connectorskillset\big)$
    \If{\texttt{Near}($\qobj^g,T$)}
        \State \textbf{Return} \texttt{Retrace}($\qobj^g, T$)
    \EndIf
\EndFor
\State \textbf{Return} None

\end{algorithmic}
\label{algo:gpuskillrrtbackbone}
\end{algorithm}
\end{minipage}
\end{center}

\subsection{Connector Policy Training}\label{Appendix:Connector}
To train the connector policy \( \pi_C\), we introduce the state space $S$, the action space $A$, and the reward function $R$ used for training. The problem setup for connector policy is provided in Section~\ref{method:train_connector}.

For training connector policy $\pi_C$, the state space, the action space, and the reward function are defined as follows. 

\begin{itemize}[leftmargin=10pt, itemsep=0.3pt]
    \item \textbf{State (\( S \)):} 

\begin{table}[H]
\centering
\begin{adjustbox}{width=0.7\textwidth} 
\begin{tabular}{|c|c|c|}
\hline
\textbf{Extract from} & \textbf{Symbol} & \textbf{Description} \\
\hline
& $q^\text{(t)}_r\in\mathbb{R}^9$ &robot joint position   \\
\cline{2-3}
\multirow{5}{*}{Robot configuration}
& $\dot{q}^\text{(t)}_r\in\mathbb{R}^9$ &Robot joint velocity  \\
\cline{2-3}
 & $T^\text{(t)}_\text{tool}\in\mathbb{R}^7$& Robot tool pose  \\
\cline{2-3}
 & $p^\text{(t)}_\text{ee}\in\mathbb{R}^{24}$& Robot end-effector keypoint positions \\
 \cline{2-3}
 & $p^\text{(t)}_\text{tip}\in\mathbb{R}^6$& Robot gripper tip positions  \\
 \cline{1-3}
 Object configuration& $p^\text{(t)}_\text{obj}\in\mathbb{R}^{24}$ &Object keypoint positions   \\
\hline
Action & $a^\text{(t-1)}\in\mathbb{R}^{21}$ & Previous action  \\
\hline
Simulator & $\mathbbm{1}^\text{(t)}_\text{gripper-execute} $ & Gripper action executability  \\
\hline
\multirow{2}{*}{Target robot configuration} & $p^\text{g}_\text{ee}\in\mathbb{R}^{24}$ &Target end-effector keypoint positions \\
\cline{2-3}
 & $p^\text{g}_\text{tip}\in\mathbb{R}^{24}$ &Target gripper tip positions  \\
\hline
\end{tabular}
\end{adjustbox}
\caption{The components of state space $S$ of connector policy $\pi_C$}\label{table:Connector_state}
\end{table}
    The state for the skill's connector policy, \( K.\pi_C \), consists of ten components outlined in Table~\ref{table:Connector_state}. 
    \(\mathbbm{1}^\text{(t)}_\text{gripper-execute}\), provided by the simulator, represents a binary indicator of the executability of the gripper action, where a value of 1 indicates that the action is executable. The gripper action is executable every 1.2 seconds because the Franka Research 3 gripper cannot accept new commands until the previous gripper command is completed. The goal is represented by the target end-effector keypoint positions \( p^\text{g}_\text{ee} \) and the target gripper tip positions \( p^\text{g}_\text{tip} \), which define the target configurations for the robot to achieve. The remaining state components are identical to the state components of the prehensile skill post-contact policy.

    \medskip

    \item \textbf{Action (\( A \)):} The action \( a \in A \) represents the control inputs applied to the robot. The action consists of four components as show in Table \ref{table:Connector_action}. The delta end-effector pose, the proportional gain for robot joints, and joint damping correspond to the action components of the NP skill's post-contact policy. The fourth component, $q_\text{width}$, is the target gripper tip width. For the robot joints (excluding the gripper tip), the control process follows the same procedure as the post-contact policy control process used in the non-prehensile skill. For the gripper tip, the gripper width is adjusted to \( q_\text{width} \).
    \begin{table}[H]
\centering
\begin{adjustbox}{width=0.4\textwidth} 
\begin{tabular}{|c|c|} 
\hline
\textbf{Symbol} & \textbf{Description}  \\
\hline
$\Delta q_\text{ee}\in\mathbb{R}^6$& \text{Delta end-effector pose}  \\ \hline
$k_p\in\mathbb{R}^7$ & \text{Proportional gain}   \\ \hline
$\rho\in\mathbb{R}^7$ & \text{Joint damping}   \\ \hline
$q_\text{width}\in\mathbb{R}$& \text{Target gripper width}  \\ \hline

\end{tabular}
\end{adjustbox}
\caption{The components of action space $A$ of connector policy $\pi_C$}\label{table:Connector_action}
\end{table}

    \medskip
    \item \textbf{Reward (\( R(s_t, a_t, s_{t+1}) \)):} 
    \begin{enumerate}
        \item \textbf{End-Effector Distance Reward:} Encourages the end-effector to move closer to its target position. Here, $p^\text{(t)}_\text{ee}$ represents the end-effector keypoint positions at timestep $t$, and $p^\text{g}_\text{ee}$ denotes the target end-effector keypoint positions:
        \[
        \begin{aligned}
        r^\text{(t)}_{\text{ee}} = \epsilon^{\text{ee}}_0 \big( &\exp(-\epsilon^{\text{ee}}_1 \| p^\text{(t)}_\text{ee} - p^\text{g}_\text{ee} \|_2) - \exp(-\epsilon^{\text{ee}}_1 \| p^\text{(t-1)}_\text{ee} - p^\text{g}_\text{ee} \|_2) \big)
        \end{aligned}
        \]
    
        \item \textbf{Gripper Tip Position Reward:} Aligns the gripper tips with their target positions. Here, $p^\text{(t)}_\text{ee}$ represents the gripper tip positions at timestep $t$, and $p^\text{g}_\text{tip}$ denotes the target tip positions:
        \[
        \begin{aligned}
        r^\text{(t)}_{\text{tip}} = \epsilon^{\text{tip}}_0 \big( &\exp(-\epsilon^{\text{tip}}_1 \| p^\text{(t)}_\text{ee} - p^\text{g}_\text{tip} \|_2)
        - \exp(-\epsilon^{\text{tip}}_1 \| p^\text{(t-1)}_\text{ee} - p^\text{g}_\text{tip} \|_2) \big)
        \end{aligned}
        \]
    
        \item \textbf{Object Movement Penalty:} Penalizes unnecessary object movement to ensure the preconditions of subsequent skills remain valid. Here, $p^\text{(t)}_\text{obj}$ represents the object keypoint positions at timesteps $t$:
        \[
        r^\text{(t)}_{\text{obj-move}} = -w^{\text{move}} \| p^\text{(t)}_\text{obj} - p^\text{(t-1)}_\text{obj} \|_2
        \]
    
        \item \textbf{Success Reward:} A success reward, \( r_{\text{succ}} \), is given when both the end-effector and gripper tip reach their respective target positions, \( p^\text{g}_\text{ee} \) and \( p^\text{g}_\text{tip} \), within the allowable error thresholds \( \delta_\text{ee} \) and \( \delta_\text{tip} \).
    \[
    r^\text{(t)}_{\text{success}} =
    \begin{cases} 
    r_{\text{succ}} & \text{if } ||p^\text{(t)}_\text{ee} - p^\text{g}_\text{ee}||_2 < \delta_\text{ee}, \text{ and }  ||p^\text{(t)}_\text{tip} - p^\text{g}_\text{tip}||_2 < \delta_\text{tip}, \\
    0 & \text{otherwise}
    \end{cases}
    \]
    \end{enumerate}

    The overall reward is defined as:
    \[
    r^\text{(t)}_{\text{connector}} = r^\text{(t)}_{\text{ee}} + r^\text{(t)}_{\text{tip}} + r^\text{(t)}_{\text{obj-move}} + r^\text{(t)}_{\text{success}}
    \]
    The hyperparameters of the reward function, $\epsilon_0$, $\epsilon_1$, $\omega$, $\epsilon_\text{vel}$, $\delta_\text{ee}$, and $\delta_\text{tip}$, vary depending on the specific connector being trained. The values of these reward hyperparameters for each connector are provided in Table~\ref{table:connector_reward}.
    
\end{itemize}

\begin{table}[H]
\centering
\begin{adjustbox}{width=0.9\columnwidth} 
\begin{tabular}{|l|c|c|c|c|c|c|c|c|c|}
\hline
\textbf{Domain} & \multicolumn{2}{c|}{\textbf{Card Flip}} & \multicolumn{3}{c|}{\textbf{Bookshelf}} & \multicolumn{4}{c|}{\textbf{Kitchen}} \\ \hline
\textbf{Skills} & slide & prehensile & topple & prehensile & push & sink & prehensile & l-cupboard & r-cupboard \\ \hline
$\epsilon^\text{ee}_0$       &  \multicolumn{9}{c|}{40} \\ \hline
$\epsilon^\text{ee}_1$       &  \multicolumn{5}{c|}{0.9}  & \multicolumn{4}{c|}{1.0} \\ \hline
$\epsilon^\text{tip}_0$       &  \multicolumn{7}{c|}{40} & \multicolumn{2}{c|}{30} \\ \hline
$\epsilon^\text{tip}_1$       & \multicolumn{9}{c|}{1.0} \\ \hline
$\omega^{\text{move}}$ &  \multicolumn{6}{c|}{-0.3} & -3.0 & \multicolumn{2}{c|}{-10} \\ \hline
$r_\text{succ}$ &   \multicolumn{5}{c|}{1000}  & \multicolumn{2}{c|}{150} & \multicolumn{2}{c|}{100} \\ \hline
$\delta_{\text{ee}}$ & \multicolumn{9}{c|}{0.01}  \\ \hline
$\delta_{\text{tip}}$ & \multicolumn{9}{c|}{0.003}  \\ \hline
\end{tabular}
\end{adjustbox}
\caption{Reward hyperparameter values for training connector policy}\label{table:connector_reward}
\end{table}

Connector policies utilize a multilayer perceptron (MLP) architecture to generate low-level robot actions based on the state and target robot configuration. Each connector policy \( \pi_C \) employs a five-layer MLP with an input dimension of 131 and an output dimension of 21. Other components of the network architecture are identical to those of the NP skill post-contact policy's network architecture.
\begin{table}[H]
    \fontsize{8}{8}\selectfont
    \centering
    
\begin{tabular}{c|c|c|c|c|c}
    \toprule
    & \makecell{input\\dimensions}& \makecell{hidden\\dimensions} & \makecell{output\\dimensions} &\makecell{hidden\\activations} &\makecell{output\\activation} \\
    \cmidrule(lr){1-1}\cmidrule(lr){2-2}\cmidrule(lr){3-3}\cmidrule(lr){4-4}\cmidrule(lr){5-5}\cmidrule(lr){6-6}
    $\pi_C$
    & 131
    & {$[512, 256, 256, 128]$}
    & 21
    & {ELU}
    & Identity
    \\
    \bottomrule
    \end{tabular}
    \label{table:connector_MLP_architecture}
    \caption{The network architecture of the connector policy}

\end{table}

\subsection{Imitation Learning Detail}

\begin{table}[h!]
\centering
\setlength\tabcolsep{20 pt}
\begin{adjustbox}{width=0.6\textwidth} 
\begin{tabular}{|c|c|}
\hline
\textbf{Symbol} & \textbf{Description} \\
\hline
$q^\text{(t)}_r \in \mathbb{R}^9$ &Robot joint position  \\
\cline{1-2}
 $q^\text{(t-1)}_r \in \mathbb{R}^9$ & Previous robot joint position  \\
\cline{1-2}
 $p^\text{(t)}_\text{ee} \in \mathbb{R}^{24}$ & \makecell{Robot end-effector keypoint positions \\ (computed from $q^\text{(t)}_r$)} \\
\cline{1-2}
 $p^\text{(t)}_\text{tip} \in \mathbb{R}^6$ & \makecell{Robot gripper tip positions \\ (computed from $q^\text{(t)}_r$)} \\
\hline
$p_\text{ee-rel}^\text{(t)} \in \mathbb{R}^{24}$ & \makecell{Relative robot end-effector \\ keypoint positions w.r.t $q^\text{(t)}_\text{obj}$} \\
\cline{1-2}
 $p^\text{(t)}_\text{tip-rel}\in \mathbb{R}^6$ & \makecell{Relative robot gripper \\ tip positions w.r.t $q^\text{(t)}_\text{obj}$} \\
\hline
  $p^\text{(t)}_\text{obj} \in \mathbb{R}^{24}$ &\makecell{Object keypoint positions \\ (computed from $q^\text{(t)}_\text{obj}$)}  \\
\hline
 $ \mathbbm{1}^\text{(t)}_\text{gripper-execute} $ & \makecell{Gripper action executability \\ (executable every 1.2 seconds)} \\
\hline
 $ a^\text{(t-1)}_\text{width}\in \mathbb{R}$ & \makecell{Previous timestep's \\ robot gripper width action} \\
\hline
$p^\text{g}_\text{obj} \in \mathbb{R}^{24}$ & Goal object keypoint positions \\
\hline
\end{tabular}
\end{adjustbox}
\begin{flushleft}
\footnotesize
\caption{The components of state space $S$ of distillation policy.}\label{table:IL_state}
\end{flushleft}
\end{table}


\begin{table}[h!]
\centering
\begin{adjustbox}{width=0.6\textwidth}
\setlength\tabcolsep{20 pt}
\begin{tabular}{|c|c|}
\hline
 \textbf{Symbol} & \textbf{Description} \\
\hline
$\Delta q_\text{joint} \in \mathbb{R}^7$ & \makecell{Changes in the robot joint positions \\ (except the two gripper joints)}  \\
\hline
 $q_\text{width} \in \mathbb{R}$ & Target gripper width  \\
\hline
 $k_p \in \mathbb{R}^{7}$ & \makecell{Proportional gain of robot joints \\ (except the two gripper joints).} \\
 \hline
 $\rho \in \mathbb{R}^{7}$ & \makecell{Damping of robot joints \\ (except the two gripper joints).} \\
\hline
\end{tabular}
\end{adjustbox}
\vspace{-0.2cm}
\begin{flushleft}
\footnotesize
\caption{The components of action space $A$ of distillation policy.}\label{table:IL_action}
\end{flushleft}
\vspace{-0.6cm}
\end{table}

In this section, we describe the diffusion policy used for imitation learning, including its architecture, input/output structure, and training configuration.

We train the diffusion policy with a U-Net backbone from the diffusion policy \cite{chi2023diffusion} codebase. To achieve faster inference times, we remove action chunking and state history.

For the state components, as shown in Table \ref{table:IL_state}, we use the previous robot joint position \( q^\text{(t-1)}_r \) instead of the robot joint velocity $\dot{q}^\text{(t)}_r$ due to the large sim-to-real gap in joint velocity. The relative positions of the robot's end-effector keypoints $p^\text{(t)}_\text{ee-rel}$ and gripper tip $p^\text{(t)}_\text{tip-rel}$, with respect to the object pose $q_\text{obj}$, are used to explicitly represent the relationship between the object and the robot. We also use \( a^\text{(t-1)}_\text{width} \) to identify the previous width action in order to provide previous gripper commands for efficient execution. Other state components (\( q^\text{(t)}_r, p^\text{(t)}_\text{ee}, p^\text{(t)}_\text{tip}, p^\text{(t)}_\text{obj}, \) and \( p^\text{g}_\text{obj} \)) correspond to the same components in the state components of the prehensile skill post-contact policy. The state component \( \mathbbm{1}^\text{(t)}_\text{gripper-execute} \) corresponds to the connector policy's state component.

For the action components, as described in Table \ref{table:IL_action}, it consists of four components to perform torque control for non-prehensile manipulation. Their gains and damping values correspond to the same components in the action components of the non-prehensile skill post-contact policy. The distillation policy uses changes in the robot joint positions $\Delta q_\text{joint}$ instead of changes in the end-effector $\Delta q_\text{ee}$, eliminating the need for inverse kinematics (IK) computation and enabling faster inference. Hyperparameters related to training diffusion policies are summarized in Table \ref{table:IL_hyper}. These parameters remain consistent across the card flip, bookshelf, and kitchen domains.
\begin{table}[H]
    \centering
    \begin{adjustbox}{width=\columnwidth} 
    \setlength\tabcolsep{9 pt}
    \begin{tabular}{cccccccccc}
    \toprule
    To & Ta & Tp & Hidden Dimension & LR & Weight Decay & Batch Size & D-Iters Train & D-Iters Eval & D-Iters Emb Dim \\
    \midrule
    1 & 1 & 1 & [256,512,1024,2048] & 1e-4 & 1e-6 & 4096 & 100 & 8 & 256 \\
    \bottomrule
    \end{tabular}
    \end{adjustbox}
    \caption{Hyperparameters of U-Net based diffusion policy. To: observation horizon, Ta: action horizon, Tp: action prediction horizon, Hidden Dimension: hidden dimension of U-Net, LR: learning rate, Weight Decay: weight decay of optimizer, Batch Size: size of mini batch in training, D-Iters Train: number of training diffusion iterations, D-Iters Eval: number of inference DDIM iterations, D-Iters Emb Dim: embedding size of diffusion timestep}
    \label{table:IL_hyper}
\end{table}
\label{Appendix:imitation_learning}

\subsection{Domain Randomization}
For sim-to-real transfer, we apply domain randomization throughout the entire pipeline, including skill training, \texttt{Skill-RRT}, and data collection. To mimic perception noise, we inject noise into the object pose, robot joint positions, and end-effector pose. We apply different scales of noise to the object pose, as the perception noise of the object pose increases depending on the environment. For example, when a book is fitted between two other books in the bookshelf domain, the noise in the object pose is larger because the book is only partially visible. We also randomize the friction of the environment since real-world friction is unknown and add random noise to the commanded torque to reflect real-world noise effects in robot controller. The noises for each domain are summarized in Table \ref{table:DR_params}.

\begin{table}[H]
\centering
\begin{adjustbox}{width=0.7\columnwidth} 
\begin{tabular}{|c|c|c|c|}
\hline
\textbf{}                   & \multicolumn{3}{c|}{\textbf{Range}}           \\ \hline
\textbf{Domain}                   & \textbf{Card Flip}     & \textbf{Bookshelf}     & \textbf{Kitchen}           \\ \hline
Object Pose (Position)              & $+ N[0.0, 0.003]$      & $+ N[0.0, 0.005]$ & $+ N[0.0, 0.003]$ \\ \hline
Object Pose (Orientation)           & $+ N[0.0, 0.03]$       & $+ N[0.0, 0.05]$ & $+ N[0.0, 0.03]$ \\ \hline
Robot joint position                & \multicolumn{3}{c|}{$+ N[0.0, 0.005]$}   \\ \hline
EE Pose (Position)                  & \multicolumn{3}{c|}{$+ N[0.0, 0.001]$}   \\ \hline
EE Pose (Orientation)               & \multicolumn{3}{c|}{$+ N[0.0, 0.01]$}    \\ \hline
Environment friction                & \multicolumn{3}{c|}{$\times U[0.8, 1.2]$}\\ \hline
Robot EE surface friction           & \multicolumn{3}{c|}{$\times U[0.9, 1.1]$}\\ \hline
Object mass                         & \multicolumn{3}{c|}{$\times U[0.8, 1.2]$}\\ \hline
Torque noise                        & \multicolumn{3}{c|}{$+ N[0.0, 0.03]$}    \\ \hline
\end{tabular}
\end{adjustbox}
\caption{Ranges of domain randomization. \( U[\text{min}, \text{max}] \) denotes uniform distribution, and \( N[\mu, \sigma] \) denotes normal distribution. The symbol "$+$" represents the summation operation, and "$\times$" represents the product operation.}\label{table:DR_params}
\end{table}
\label{Appendix:DR}

\newpage
\section{Detail of Main Experiments}\label{Appendix:Main_Exp}
In this section, we present (1) the detailed training setup for the baselines used in our main experiment (Section~\ref{sec:main_exp} Table~\ref{table:main_exp}), (2) an in-depth analysis of the main experimental results, and (3) the setup for the real-world experiments in Section~\ref{sec:real_exp}.

\subsection{Detail of Baselines}
\subsubsection{Baseline PPO Detail}\label{Appendix:baseline_PPO}

Baseline PPO is a flat reinforcement learning method whose training pipeline follows the same process as skill training. To train the policy \( \pi_\text{post} \), we detail: (1) the initial problem setup, (2) the state space \( S \), the action space \( A \), and the reward function \( R \), and (3) a policy architecture used for training.

The problem for training the post-contact policy, \(\pi_\text{post} \), consists of: (1) the initial object pose \( q^\text{init}_\text{obj} \), (2) the initial robot configuration \( q^\text{init}_r \), and (3) the target object pose \( q^\text{g}_\text{obj} \). We randomly sample the initial and desired object poses, the same as in Skill-RRT in Appendix~\ref{Appendix:Problem}.
Subsequently, we compute \( q^\text{init}_r \) by \( \pi_\text{pre}(q^\text{init}_\text{obj},q^\text{g}_\text{obj}) \). If the computed \( q^\text{init}_r \) does not cause collisions, either between the robot and the environment or between the robot and the object, the problem is generated. Otherwise, it is excluded from the dataset.

\begin{itemize}[leftmargin=10pt, itemsep=0.3pt]
    \medskip
    \item \textbf{State (\( S \)):} The components of the state space for the PPO policy are identical to those of the distillation policy. However, the PPO policy uses the robot’s joint velocity $\dot{q}^\text{(t)}_r$ instead of the previous joint position $q^\text{(t-1)}_r$ and excludes the gripper action executability $\mathbbm{1}_\text{gripper-execute}$. Additionally, it incorporates not only the previous gripper width action $ a^\text{(t-1)}_\text{width}$ but also all previous actions $a^\text{(t-1)}$ as components of the state space.
    \medskip
    \item \textbf{Action (\( A \)):}
    The action components are identical to action components of connector policy. 

    \medskip
    \item \textbf{Reward (\( R \)):} The reward function \( R(s_t, a_t, s_{t+1}) \) is designed to encourage the robot to manipulate object to the goal object pose \( q_{\text{obj}}^\text{g} \). The reward consists of three main components:
    \begin{itemize}

        \item \textbf{Object Keypoint Distance Reward:} The reward $r^\text{(t)}_\text{obj}$ computation is identical to the object keypoint distance reward used in non-prehensile skill post-contact policy training. 

        \item \textbf{Tip Contact Reward:} The reward $r^\text{(t)}_\text{tip-contact}$ computation is identical to the tip contact reward used in non-prehensile post-contact policy training. 

        \item \textbf{Domain-Oriented Reward:} Encourages the successful completion of domain-specific objectives. The exact reward varies depending on the domain as shown in Table \ref{table:PPO_reward_condition}:

        \[
        \begin{aligned}
        r^\text{(t)}_{\text{domain}} = {\epsilon_0^{\text{domain}}} \cdot \mathbb{I}[\text{domain-specific conditions}]
        \end{aligned}
        \]

        \begin{table}[H]
\centering
\begin{adjustbox}{width=0.4\textwidth} 
\begin{tabular}{|c|c|}
\hline
\textbf{Domain} & \multicolumn{1}{c|}{domain-specific conditions} \\ \hline
\textbf{Card Flip} & \multicolumn{1}{c|}{The card is flipped} \\ \hline
\textbf{Bookshelf} & \multicolumn{1}{c|}{The book is placed on the box} \\ \hline
\textbf{Kitchen} & \multicolumn{1}{c|}{The cup is placed on the shelf} \\ \hline
\end{tabular}
\end{adjustbox}
\caption{domain-specific conditions for training the Baseline PPO policy}\label{table:PPO_reward_condition}
\end{table}



        \item \textbf{Success Reward:} The reward $r^\text{(t)}_\text{success}$ computation is identical to the success reward used in non-prehensile post-contact policy training.
    \end{itemize}

    The overall reward is defined as:
    \[
    r^\text{(t)}_{\text{PPO}} = r^\text{(t)}_{\text{obj}} + r^\text{(t)}_{\text{tip-obj}} + r^\text{(t)}_{\text{domain}} + r^\text{(t)}_{\text{success}}
    \]
    The hyperparameters of the reward function, $\epsilon_0^{\text{obj}}$, $\epsilon_1^{\text{obj}}$, $\epsilon_0^{\text{tip-obj}}$, $\epsilon_1^{\text{tip-obj}}$, $\epsilon_0^{\text{domain}}$ and $r_{\text{succ}}$, are provided in Table~\ref{table:baseline_ppo_reward}.
    \begin{table}[H]
\centering
\begin{adjustbox}{width=0.4\textwidth} 
\begin{tabular}{|c|c|c|c|}
\hline
\textbf{Domain} & \multicolumn{1}{c|}{\textbf{Card Flip}} & \multicolumn{1}{c|}{\textbf{Bookshelf}} & \multicolumn{1}{c|}{\textbf{Kitchen}} \\ \hline
$\epsilon_0^{\text{obj}}$       & \multicolumn{3}{c|}{0.02} \\ \hline
$\epsilon_1^{\text{obj}}$       & \multicolumn{3}{c|}{0.02} \\ \hline
$\epsilon_0^{\text{tip-obj}}$       & \multicolumn{3}{c|}{0.03} \\ \hline
$\epsilon_1^{\text{tip-obj}}$       & \multicolumn{3}{c|}{0.03} \\ \hline
$\epsilon_0^{\text{domain}}$       & \multicolumn{3}{c|}{0.5} \\ \hline
$r_{\text{succ}}$ & \multicolumn{3}{c|}{1000} \\ \hline
$\delta_\text{obj}$ & \multicolumn{3}{c|}{0.005} \\ \hline
\end{tabular}
\end{adjustbox}
\caption{Reward hyperparameter values for training the Baseline PPO policy}\label{table:baseline_ppo_reward}
\end{table}
\end{itemize}

The baseline PPO policies utilize a multilayer perceptron (MLP) architecture to generate low-level robot actions based on the state and goal object pose information. Each policy, $\pi_{\text{pre}}$ and $\pi_{\text{post}}$, employs a five-layer MLP. $\pi_{\text{pre}}$ and $\pi_{\text{post}}$ have input dimensions of 14 and 147, respectively. $\pi_{\text{pre}}$ and $\pi_{\text{post}}$ have output dimensions of 9 and 21, respectively. The other components of the architecture are identical to those of the non-prehensile skill post-contact policy's network architecture.
\begin{table}[H]
    \fontsize{8}{8}\selectfont
    \centering
\begin{tabular}{c|c|c|c|c|c}
    \toprule
    & \makecell{input\\dimensions}& \makecell{hidden\\dimensions} & \makecell{output\\dimensions} &\makecell{hidden\\activations} &\makecell{output\\activation} \\
    \cmidrule(lr){1-1}\cmidrule(lr){2-2}\cmidrule(lr){3-3}\cmidrule(lr){4-4}\cmidrule(lr){5-5}\cmidrule(lr){6-6}
    $\pi_\text{pre}$
    & 14
    & \multirow{2}{*}{$[512, 256, 256, 128]$}
    & 9 
    &
    \multirow{2}{*}{ELU} & 
    \multirow{2}{*}{Identity}
    \\
    \cmidrule(lr){1-1}\cmidrule(lr){2-2}\cmidrule(lr){4-4}
    $\pi_{\text{post}}$
    & 147
    & 
    & 21
    &
    &
    \\
    \bottomrule
    \end{tabular}
    \caption{The network architecture of the Baseline PPO policies}\label{table:baseline_ppo_network_architecture}
\end{table}

\subsubsection{Baseline \texttt{MAPLE} Detail}\label{Appendix:baseline_MAPLE}
Baseline \texttt{MAPLE} is a hierarchical reinforcement learning method that includes the task policy \( \pi_\text{tsk} \) and the parameter policy \( \pi_\text{parameter} \). The algorithm is implemented based on Soft Actor-Critic~\cite{haarnoja2018soft}, consistent with the original implementation. However, we modify the \texttt{MAPLE} implementation to incorporate a parallelized environment, Isaac Gym, and to adapt \texttt{MAPLE} to our PNP problem. First, we remove the affordance reward, which in \texttt{MAPLE} guides the high-level policy toward the desired manipulation region. Instead, we integrate the connectors and skills, incorporating an applicability checker \(\applicabilitychecker\). The connector, which moves the robot to a state where the corresponding skill is applicable, is executed only when the predicted desired object pose is applicable (i.e., when the corresponding \(\applicabilitychecker\) holds true); otherwise, it is not executed. This replaces the need for an affordance reward. Furthermore, we replace the explicit initial end-effector position parameter, \( x_{\text{reach}} \), which serves as an input to skills in \texttt{MAPLE}, with the output of the pre-contact policy \( \pi_\text{pre} \) of each skill. Therefore, the initial robot position is computed by \( \pi_\text{pre} \), and the connector policy moves the robot to this computed position before skill execution. Additionally, we eliminate atomic primitives, low-level actions used to fill in gaps that cannot be fulfilled by skills, since our connectors are already trained to handle these gaps.

To train the policies, we outline: (1) the initial problem setup, (2) the state space \( S \), the action space \( A \), and the reward function \( R \) and, (3) hyperparameters used for training. 

The problem for training the policies, \( \pi_\text{tsk} \) and \( \pi_\text{parameter} \), consists of: (1) the initial object pose \( q^\text{init}_\text{obj} \), and (2) the target object pose \( q^\text{g}_\text{obj} \). We randomly sample the initial and target object poses, the same as in Skill-RRT in Appendix~\ref{Appendix:Problem}.

\begin{itemize}[leftmargin=10pt, itemsep=0.3pt]
    \medskip
    \item \textbf{State (\( S \)):} The components of the state space for the \texttt{MAPLE} policies are identical to those of the distillation policy. However, the \texttt{MAPLE} policies utilize the robot’s joint velocity $\dot{q}^\text{(t)}_r$ instead of the previous joint position $q^\text{(t-1)}_r$ and omits both the gripper action executability $\mathbbm{1}_\text{gripper-execute}$ and previous timestep's robot gripper width action $a^\text{(t-1)}_\text{width}$.

    \medskip
    \item \textbf{Action (\( A \)):} The action \( (K, \hat{q}^{\text{obj}}_{\text{sg}}) \in A \) represents skills $K$ and its corresponding normalized desired object pose $\hat{q}^{\text{obj}}_{\text{sg}}$ for the control module \( \pi_{\text{skill}}(\cdot;q^{\text{obj}}_{\text{sg}})\). The normalized desired object pose is unnormalized to align with the region of the corresponding skill.
    
    \medskip
    
    \item \textbf{Reward (\( R \)):} The reward function \( R(s, K, \hat{q}^{\text{obj}}_{\text{sg}}, s') \) is designed as a sparse reward to encourage the robot to manipulate the object to the goal object pose \( q^{\text{obj}}_\text{g} \), as defining a dense reward for this task is challenging. The reward consists of :

    \begin{itemize}

        \item \textbf{Feasible Skill Reward:} Encourages the policy to choose feasible skills.
        \[
        \begin{aligned}
        r^\text{(t)}_{\text{feasible}} = {\epsilon_0^{\text{feasible}}} \cdot \mathbbm{1}[K.\phi (s, q^{\text{obj}}_{\text{sg}}) ]
        \end{aligned}
        \]
        
        \item \textbf{Success Reward:} The reward $r^\text{(t)}_\text{success}$ computation is identical to the success reward used in non-prehensile post-contact policy training. 
    \end{itemize}

    The overall reward is defined as:
    \[
    r^\text{(t)}_{\texttt{MAPLE}} =  r^\text{(t)}_{\text{feasible}} + r^\text{(t)}_{\text{success}}
    \]
    The hyperparameters of the reward function, $\epsilon_0^{\text{feasible}} = 0.1$, $\delta_\text{obj} = 0.005$ and $r_{\text{succ}} = 100$, are the same for all tasks.
\end{itemize}

\begin{table}[h!]
\centering
\begin{adjustbox}{width=0.6\textwidth} 
\begin{tabular}{|c|c|}
\hline
\textbf{Hyperparameter} & \textbf{Value} \\ \hline
Hidden sizes (all networks) & 1024, 512, 256, 256 \\ \hline
Q network and policy activation & ReLU \\ 
Q network output activation & None \\ 
Policy network output activation & tanh \\ \hline
Optimizer & Adam \\ 
Batch Size & 4096 \\ 
Learning rate (all networks) & 3e-5 \\ 
Target network update rate \( \tau \) & 5e-4 \\ \hline
\# Training steps per epoch & 1000 \\ 
Replay buffer size & 1e6 \\ 
 \hline
Discount factor & 0.99 \\ 
Reward scale & 100.0 \\ 
Automatic entropy tuning & True \\ 
Target Task Policy Entropy & \( 0.50 \times \log(k), k \text{ is number of skills} \) \\ 
Target Parameter Policy Entropy & \( -\max_a d_a \) \\ \hline
\end{tabular}
\end{adjustbox}
\caption{Hyperparameters for the baseline \texttt{MAPLE}}\label{tab:baseline_maple_hyperparams}
\end{table}
Hyperparameters related to training \texttt{MAPLE} are summarized in Table ~\ref{tab:baseline_maple_hyperparams}. These parameters remain consistent across all the tasks. 

\textbf{Additional Analysis} \texttt{MAPLE} struggles in Card Flip and Kitchen domains due to the narrow passage problem~\cite{hsu1998finding}. For instance, in the Card Flip domain, the card can only be flipped near the edges of the table, which constitute a very small region in $\Qobj$. Moreover, even at the edge, to prevent the card from dropping during the transition from sliding to the prehensile skill, the card must be positioned precisely: it must be graspable while remaining sufficiently close to the table surface to avoid falling during the transition Similarly, in the Kitchen domain, the robot must position the cup such that the prehensile skill can be applied following the non-prehensile skill in the sink. The set of cup poses suitable for applying the prehensile skill is very narrow: the pose must be collision-free, and a valid inverse kinematics solution must exist. As a result, most attempts at exploring object subgoal poses lead to failure for \texttt{MAPLE}, which ends up learning only locally optimal behaviors, such as bringing the card to the edge of the table but avoiding the prehensile action to prevent the card from falling. In the Bookshelf domain, where the robot simply needs to topple the book to enable grasping, \texttt{MAPLE} achieves an 83\% success rate.

\subsubsection{Baseline \texttt{HLPS} Detail}\label{Appendix:baseline_HLPS}
\texttt{HLPS}\cite{wang2024probabilistic} is a hierarchical reinforcement learning method in which the high-level policy selects latent subgoals and the low-level policy produces actions to achieve them. The algorithm is based on Soft Actor-Critic\cite{haarnoja2018soft}, and we follow the original implementation with necessary modifications. Specifically, we adapt HLPS to support parallelized environments using Isaac Gym.

The state space for both the high-level and low-level policies is identical to that of the \texttt{PPO} baseline. However, the low-level policy additionally receives the latent subgoal sampled from the high-level policy as part of its input. The action space of the low-level policy is identical to that of the \texttt{PPO} baseline, where it outputs joint torque commands along with joint gain and damping values.

Both the high-level and low-level policies are trained with dense rewards. The high-level policy uses the same dense task reward as the PPO baseline, based on the distance between the current object pose and the target object pose. The low-level policy is trained with dense subgoal rewards, defined by the distance between the current state and the latent subgoal output by the high-level policy.

A key feature of HLPS is its latent subgoal representation. Instead of selecting subgoals directly in the object pose space, the high-level policy selects subgoals in a learned latent space structured by a probabilistic model using a Gaussian Process (GP). These latent subgoals are combined with the current state representation and provided to the low-level policy, which is trained to reach them.

Hyperparameters related to training \texttt{HLPS} are summarized in Table~\ref{tab:baseline_hlps_hyperparams}, and remain consistent across all tasks.

\begin{table}[h!]
\centering
\begin{adjustbox}{width=0.6\textwidth} 
\begin{tabular}{|c|c|}
\hline
\textbf{Hyperparameter} & \textbf{Value} \\ \hline
\multicolumn{2}{|c|}{\textbf{Network Architecture}} \\ \hline
Actor network hidden sizes & 400, 400, 400, 400 \\ 
Critic network hidden sizes & 400, 400, 400, 400, 400, 400 \\ 
Encoding network hidden size & 100 \\ 
Hidden layer activation (all networks) & ReLU \\ 
Actor output activation & Tanh \\ 
Optimizer & Adam \\ \hline
\multicolumn{2}{|c|}{\textbf{Training Parameters}} \\ \hline
Latent GP learning rate & \( 1 \times 10^{-5} \) \\ 
Latent GP update frequency & 10 \\ 
Batch GP time window size & 3 \\ 
Latent subgoal dimension & 7 \\ 
Learning rate (actor/critic, both levels) & 0.0002 \\ 
High-level action interval \( k \) & 50 \\ 
Target network smoothing coefficient & 0.005 \\  
Discount factor \( \gamma \) & 0.99 \\ 
Encoding layer learning rate & 0.0001 \\ 
\hline
\end{tabular}
\end{adjustbox}
\caption{Hyperparameters for the baseline \texttt{HLPS}}\label{tab:baseline_hlps_hyperparams}
\end{table}

\subsection{Detailed Results of Simulation Experiments}
The table~\ref{table:data_summary} compares the number of state-action pairs (in billions) used for training across methods and domains. Notably, SPIN requires orders of magnitude fewer data samples than traditional RL-based methods like PPO, HLPS, and MAPLE, demonstrating its data efficiency. Skill-RRT is excluded from training as it is a non-learning baseline.

Table~\ref{table:main_exp_detail} provides a detailed version of the table~\ref{table:main_exp} presented in Section~\ref{sec:Experiments}, reporting the average success rate and computation time across three seeds on 100 test problems per domain. \texttt{SPIN} achieves the highest success rate across all domains, significantly outperforming other baselines, while maintaining reasonable computation time comparable to \texttt{SPIN}-w/o-\texttt{filtering}. In contrast, purely RL-based methods (PPO, \texttt{HLPS}) show zero success, highlighting the advantage of combining planning and learning.

\begin{table}[h]
\centering
\begin{adjustbox}{width=0.5\textwidth} 
\begin{tabular}{l|c|c|c}
\toprule
Method & Card Flip & Bookshelf & Kitchen \\
\midrule
PPO             & 2.5        & 2.5        & 2.5        \\
\texttt{HLPS}           & 0.15      & 0.23      & 0.28     \\
\texttt{MAPLE}           & 0.27      & 0.33      & 0.27      \\
\texttt{SPIN}-w/o-filtering  &  0.0030  &   0.0028      & 0.0031        \\
\texttt{SPIN}      & 0.0029    & 0.0028    & 0.0032    \\
\bottomrule
\end{tabular}
\end{adjustbox}
\caption{Number of state-action pairs used for training across different methods and domains (in billions). Note that \texttt{Skill-RRT} does not require training, as it simply replays the first found skill plan in the simulator.}
\label{table:data_summary}
\end{table}

\begin{table*}[ht]

\begin{adjustbox}{width=\columnwidth} 

\centering
\begin{tabular}{cccc|ccccccc}
    \toprule
    & \multicolumn{3}{c}{Components} & \multicolumn{6}{c}{Problem Domain} \\
    \cmidrule(lr){2-4} \cmidrule(lr){5-10}
    & \multirow{2}{*}{Method}
    & \multirow{3}{*}{\makecell{Action \\ Type}} & \multirow{3}{*}{Use $\mathcal{K}$} &
    \multicolumn{2}{c}{Card Flip} & \multicolumn{2}{c}{Bookshelf} & \multicolumn{2}{c}{Kitchen}  \\
    & & &  &
    \makecell{Success \\ rate (\%)} & \makecell{Computation \\ time (s)} &\makecell{Success \\ rate (\%)} & \makecell{Computation \\ time (s)} &\makecell{Success \\ rate (\%)} & \makecell{Computation \\ time (s)}  \\
    \cmidrule(lr){1-1} \cmidrule(lr){2-4} \cmidrule(lr){5-6}\cmidrule(lr){7-8}\cmidrule(lr){9-10}
    PPO \cite{schulman2017proximal} & 
    \textcolor{orange}{Flat RL} & 
    \textcolor{blue}{Low-level action} & $\textcolor{red}{\xmark}$ &
    {\makecell{$0.0\pm0.0$}} & N/A
    &
    {\makecell{$0.0\pm0.0$}} & N/A
    &
    {\makecell{$0.0\pm0.0$}} & N/A
    \\
    \midrule
    \texttt{HLPS} \cite{wang2024probabilistic} & 
    \textcolor{red}{Hierarchical RL} & 
    \textcolor{blue}{Low-level action} & $\textcolor{red}{\xmark}$  &
    {\makecell{$0.0\pm0.0$}} &
    N/A &
    {\makecell{$0.0\pm0.0$}} &
    N/A &
    {\makecell{$0.0\pm0.0$}} & 
    N/A
    \\
    \midrule
    \texttt{MAPLE} \cite{nasiriany2022augmenting} & 
    \textcolor{red}{Hierarchical RL} & 
    \textcolor{red}{Skill $\&$ Parameter} &  $\textcolor{codegreen}{\cmark}$  &
    {\makecell{$0.0\pm0.0$}} &
    N/A &
    {\makecell{$78.0\pm2.4$}} &
    $5.3\pm2.7$&
    {\makecell{$0.0\pm0.0$}} & 
    N/A
    \\
    \midrule
    \texttt{Skill-RRT} & 
    \textcolor{darkgreen}{Planning} & 
    \textcolor{red}{Skill $\&$ Parameter} &  $\textcolor{codegreen}{\cmark}$  &
    $39.0 \pm 0.0 $ & $85.3 \pm 48.7$
    & $66.0 \pm 0.0 $
    & $79.2 \pm 67.1$
    & $64.0 \pm 0.0 $
    & $121 \pm 39.5$
    \\
    \midrule
    \texttt{SPIN}-w/o-\texttt{filtering} & 
    \textcolor{blue}{Planner distilled via IL} & 
    \textcolor{blue}{Low-level action} & $\textcolor{codegreen}{\cmark}$   &
    {\makecell{$82.0 \pm 0.0$}} & 
    \makecell{$2.68 \pm 0.61$}
    &
    {\makecell{$83.0 \pm 0.5$}} &
    \makecell{$2.93 \pm 2.05$}
    &
    {\makecell{$87.0 \pm 0.8$}} &
    \makecell{$3.02 \pm 0.65$}
    \\
    \midrule

    \texttt{SPIN} & 
    \textcolor{blue}{Planner distilled via IL} & 
    \textcolor{blue}{Low-level action} & $\textcolor{codegreen}{\cmark}$   &
    {\makecell{$95.0 \pm 0.5$}} & 
    \makecell{$2.68 \pm 0.61$}
    &
    {\makecell{$93.0 \pm 1.4$}} &
    \makecell{$2.93 \pm 2.05$}
    &
    {\makecell{$98.0 \pm 0.5$}} &
    \makecell{$3.02 \pm 0.65$}
    \\
    \bottomrule
\end{tabular}
\end{adjustbox}

\caption{Comparison of baselines based on their components (method, action type, and whether use $\mathcal{K}$ or not) and performance metrics (success rate and computation time) with their average and standard deviation across three different seeds for each problem domain. The success rate is measured on a set of 100 test problems. Computation time refers to the average elapsed time required to solve the given 100 problems when the method successfully completes them.}
\label{table:main_exp_detail}
\end{table*}
\label{Appendix:detail_main_exp}

\subsection{Real-World Experiments Detail}
In this section, we present a detailed setup for real-world experiments, covering the environment configuration and perception systems for three domains: card flip, bookshelf, and kitchen.

\begin{figure*}[ht!]
\centering
\resizebox{\textwidth}{!}{
    \includegraphics{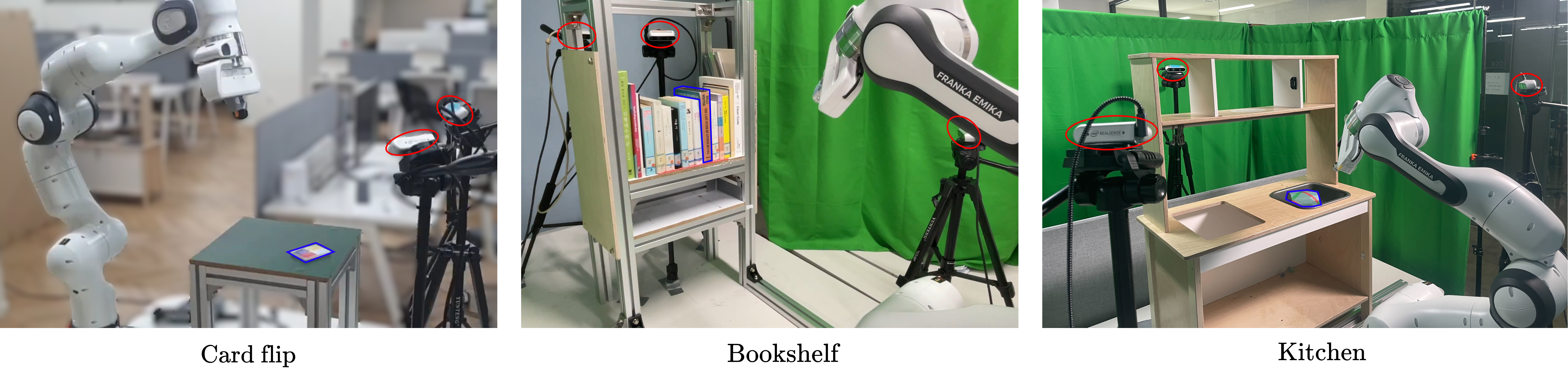}
}
\caption{The real-world setup for each domain is illustrated, with blue polygons representing the target objects and red circles indicating the camera locations.}\label{fig:real_setup}
\end{figure*}

\textbf{Environment Configuration} We utilize the Franka Research 3 robot and its gripper across three domains. To enhance the gripper's effectiveness in non-prehensile manipulation, rubber is added to the gripper fingers to increase surface friction. The environment setups are illustrated in Figure \ref{fig:real_setup}.

\begin{itemize}
    \item \textbf{Card Flip} For the real-world card flip domain, we construct a 30x30x40 cm table identical to the simulation table. The 5x7x0.5 cm card (cuboid shape) is textured with colors to break symmetry in its shape.
    
    \item \textbf{Bookshelf} In the real-world bookshelf domain, we build a bookshelf similar to the one used in simulation and utilize a 14x20x3 cm book (cuboid shape). To facilitate book toppling, sandpaper is affixed to the top surface of the book to enhance shear contact force.
    
    \item \textbf{Kitchen} The kitchen domain utilizes the IKEA DUKTIG play kitchen and a cup with a body that cannot be grasped. To enhance grasping stability, the cup handle is filled with foamboard.
\end{itemize}

\textbf{Perception System} The perception modules are used to estimate object poses in the real world. To address potential occlusions caused by the robot, multiple RealSense D435 cameras are installed, as shown in Figure \ref{fig:real_setup}. For object pose estimation, we use FoundationPose \cite{ornek2025foundpose}, an off-the-shelf RGB-D-based object pose estimator. During each episode, we select the camera with the best visibility, determined by the largest object segmentation mask. The object segmentation mask is initially generated using SAM \cite{kirillov2023segment} and subsequently updated over time using Cutie \cite{cheng2024putting}.

\label{Appendix:real_exp}

\newpage
\section{Ablation Studies}\label{Appendix:Ablation}
In this section, we provide a detailed description of the ablation studies conducted to validate the design choices of our framework. Specifically, we investigate three aspects: (1) the necessity of learning the \textit{connector} policy via reinforcement learning, by comparing it with a motion planner; (2) the effectiveness and importance of our \textit{data filtering} method during imitation learning; and (3) the impact of different \textit{model architectures} on imitation learning performance.

\subsection{Connector Effectiveness Ablations}\label{Appendix:Connector_vs_MP}
We need connector policies that bridge the state gaps between the termination of one skill and the initiation of the next. We use RL to train the connector policies, while motion planners (MP) could be used. There are two main reasons: (1) RL policies achieve higher success rates in bridging these gaps, especially during contact and de-contact between the robot and the object, and (2) neural network-based policies offer fast inference times, which help accelerate our planner, \texttt{Skill-RRT}. To demonstrate the necessity and robustness of our connectors, we compare them with MP, specifically Bi-RRT from OMPL~\cite{sucan2012the-open-motion-planning-library}.

The success rate of the MP is lower than that of the connectors in almost all cases, as shown in Table~\ref{table:c_vs_mp}. In the Kitchen domain, Bi-RRT fails to find a valid joint trajectory from the non-prehensile's terminal state to the prehensile's initial state because the narrow sink space causes frequent collisions between the robot and the environment or the object. Additionally, our connectors offer significantly shorter inference times compared to the motion planner, which requires replanning for every problem instance.
\begin{table}[h]
\centering
\begin{adjustbox}{width=0.7\textwidth} 
\begin{tabular}{c|cc|cc|cc}
\toprule
\multirow{1}{*}{Method} & \multicolumn{2}{c|}{Card} & \multicolumn{2}{c|}{Bookshelf} & \multicolumn{2}{c}{Kitchen} \\
                        & \multicolumn{1}{c}{NP $\rightarrow$ P} & \multicolumn{1}{c}{P $\rightarrow$ NP} & \multicolumn{1}{c}{NP $\rightarrow$ P} & \multicolumn{1}{c}{P $\rightarrow$ NP} & \multicolumn{1}{c}{NP $\rightarrow$ P} & \multicolumn{1}{c}{P $\rightarrow$ NP} \\
\midrule
\midrule
Connectors & 91.0 & 88.0 & 99.0 & 94.0 & 90.0 & 78.0 \\
Bi-RRT & 80.0 & 86.0 & 55.0 & 61.0 & 0.0 & 81.0 \\
\bottomrule
\end{tabular}
\end{adjustbox}
\caption{Comparison of the success rates between connectors and Bi-RRT on 100 connector tasks in simulation. Success indicates success rate among 100 problems with unit (\%).}
\label{table:c_vs_mp}
\end{table}

\begin{figure}[h]
\centering
\includegraphics[width=\columnwidth]{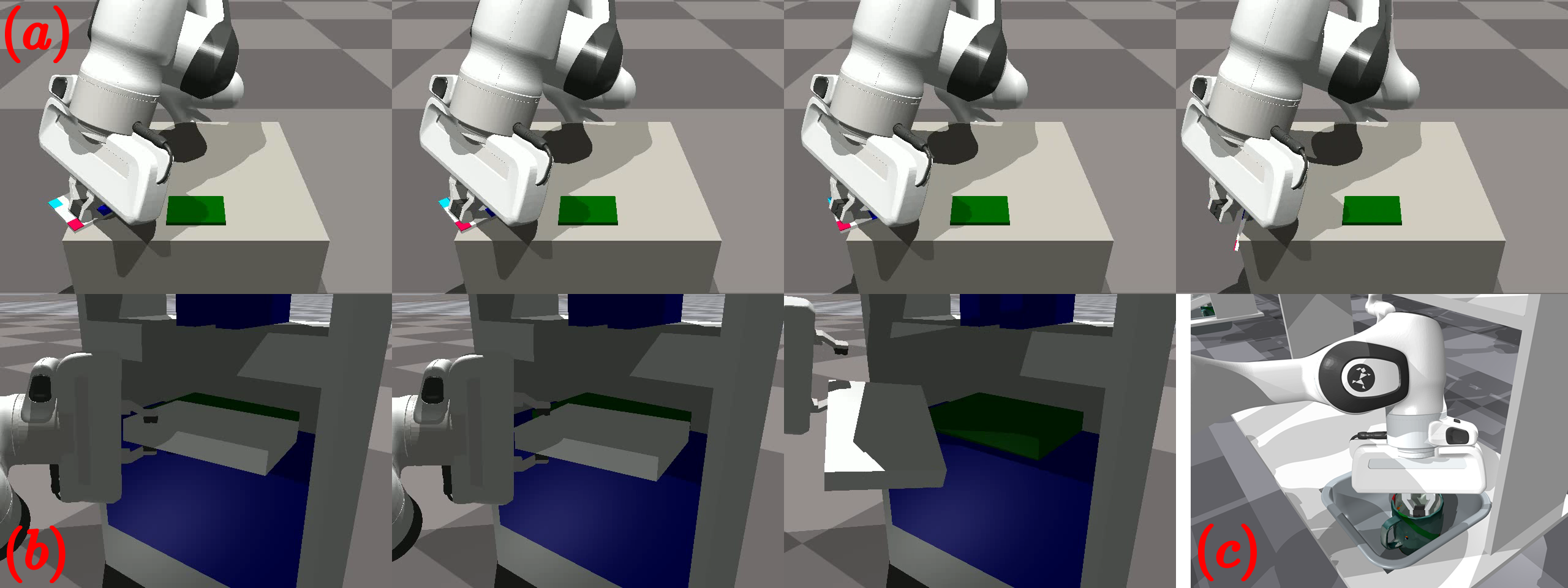}
\caption{Illustration of motion planner failure cases.}\label{fig:mp_failure}
\end{figure}

As illustrated in Fig. \ref{fig:mp_failure}, failures occur in the following scenarios: (a) the robot cannot exactly follow the planned path, leading to the card being dropped during de-contact, (b) the object is not in a perfectly stable pose while the MP assumes a static scene, causing continued contact with the gripper even after the release, or (c) the robot is surrounded by obstacles at the initial state, and it becomes computationally expensive to sample a collision-free path. These cases motivate us to train connectors using RL. We strongly encourage referring to the supplementary video (MP), which illustrates these failure cases in action.

\subsection{Filtering Method}
In this section, we study the impact of the replay success rate threshold parameter \( m \) and the effectiveness of our data filtering mechanism. As emphasized by~\citet{mandlekar2022matters}, the quality of the training dataset plays a critical role in the performance of imitation learning policies. We show that our filtering method significantly enhances data quality, leading to improved policy performance.

For the baseline (\textit{Without Replay}), we collect 15,000 skill plans without applying any filtering, using one trajectory per skill plan to train the diffusion policy. In contrast, \textit{Replay with \( m=0.1 \)} and \textit{Replay with \( m=0.9 \)} filter out skill plans that do not meet their respective replay success rate thresholds. The replay success rate of a skill plan is defined as
\[
\frac{\text{Number of successful replays of }\skillplan: N_{success}}{\text{Total number of replays of }\skillplan: N}
\].
A skill plan is included in \textit{Replay with \( m=0.1 \)} if \( N_{\text{success}} > 40 \), and in \textit{Replay with \( m=0.9 \)} if \( N_{\text{success}} > 360 \).

For both replay settings, we collect 500 skill plans. From each skill plan, we randomly sample 30 successful trajectories, resulting in a total of 15,000 trajectories per method to ensure a fair comparison during training. After collecting the trajectories, we train the distillation policy (Diffusion Policy~\cite{chi2023diffusion}) using the datasets filtered by each method via IL. The training and evaluation procedures are identical to those described in Appendix~\ref{Appendix:IL_arch_ablation}.

Table~\ref{table:ablation_dataQ} compares three different filtering methods: \textit{Without Replay}, \textit{Replay with \( m = 0.1 \)}, and \textit{Replay with \( m = 0.9 \)} (ours). The table summarizes key metrics, including the number of skill plans collected, the number of trajectories generated per skill plan, and the success rates achieved across different domains. As shown, our method (\textit{Replay with \( m = 0.9 \)}) consistently achieves the highest success rates.

\begin{table*}[ht]
\centering
\begin{adjustbox}{width=0.7\textwidth} 
\begin{tabular}{ccc|cc|c|c|c}
    \toprule
    &  & \multicolumn{2}{c}{} & & \multicolumn{3}{c}{Domain name}\\
    \cmidrule(lr){6-8}
    & \multicolumn{2}{c|}{Characteristic} & \multicolumn{2}{c|}{Data Collection} & Card Flip & Bookshelf & Kitchen   \\
    \cmidrule(lr){2-5} \cmidrule(lr){6-6}\cmidrule(lr){7-7}\cmidrule(lr){8-8}
    & Replay & $m=0.9$ & \makecell{Skill \\ plans} & \makecell{Traj \\ per a plan} &
    \makecell{Success \\ rate (\%)}  &  \makecell{Success \\ rate (\%)}  &\makecell{Success \\ rate (\%)}  \\
    \cmidrule(lr){1-1} \cmidrule(lr){2-5} \cmidrule(lr){6-6}\cmidrule(lr){7-7}\cmidrule(lr){8-8}
    \makecell{\textit{Without}\\ \textit{Replay}} & 
    $\textcolor{red}{\xmark}$ & 
    $\textcolor{red}{\xmark}$
    & 15000
    & 1
    & 79
    & 95
    & 86
    \\
    \midrule
    \makecell{\textit{Replay} \\ \textit{with $(m=0.1)$}} & 
    $\textcolor{blue}{\cmark}$ & 
    $\textcolor{red}{\xmark}$ 
    & 500
    & 30
    & 82
    & 83
    & 87
    \\
    \midrule
    \makecell{Ours (\textit{Replay}\\ \textit{with $(m=0.9)$})} & 
    $\textcolor{blue}{\cmark}$ & 
    $\textcolor{blue}{\cmark}$
    & 500
    & 30
    & \textbf{94}
    & \textbf{96}
    & \textbf{98}
    \\
    \bottomrule
\end{tabular}
\end{adjustbox}

\caption{Comparison of three different filtering methods: \textit{Without Replay}, \textit{Replay with \( m = 0.1 \)}, and Ours (\textit{Replay with \( m = 0.9 \)}). The table shows several metrics, including the characteristics, the number of skill plans collected, the number of trajectories used for training from each skill plan, and the success rates for each domain across the three filtering methods. "Skill plans" refers to the number of skill plans for each filtering method. "Traj per a plan" refers to the number of successful trajectories used for training from each skill plan. "Success rate" refers to the task success rate of the distillation policy, which is trained with data from each filtering method.}

\label{table:ablation_dataQ}
\end{table*}

\begin{figure}[h!] 
\centering
\includegraphics[width=0.8\columnwidth]{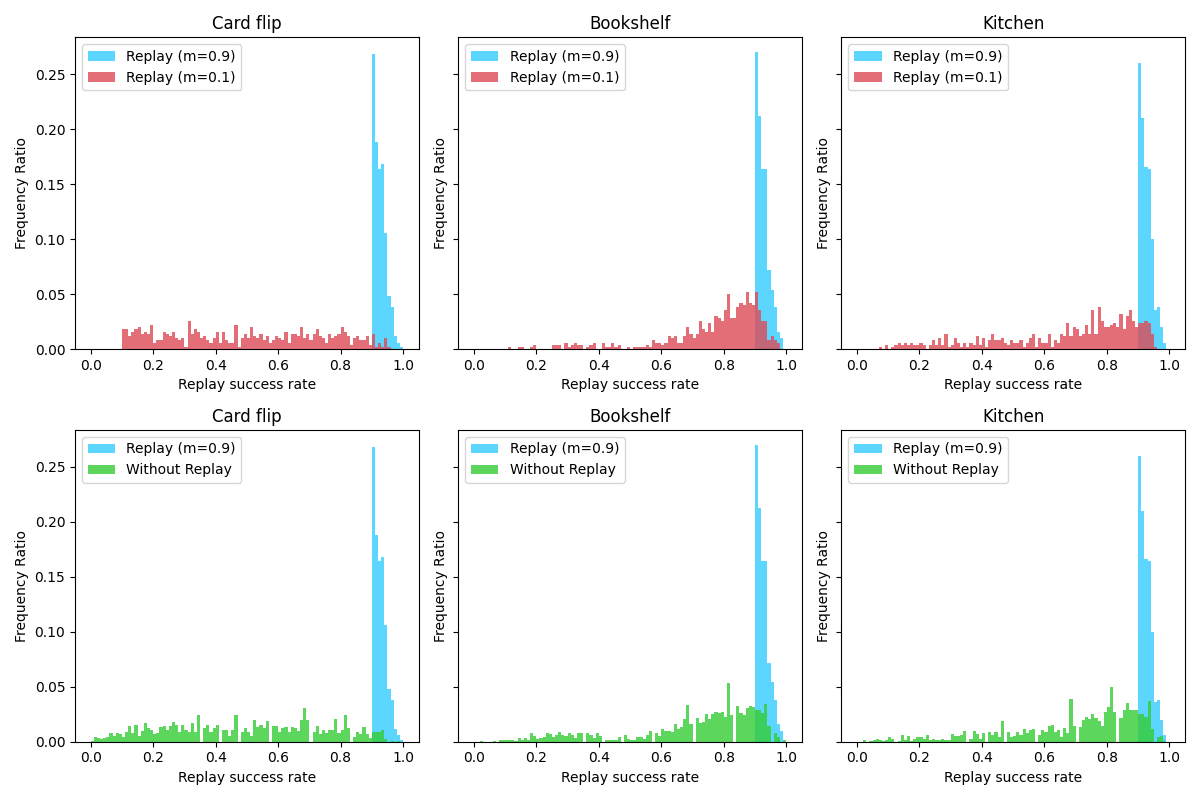}
\caption{Compare collected skill plan's replay success rate for each filtering method} 
\label{fig:replay_success_dataQ} 
\end{figure}

\begin{figure}[h] 
\centering
\includegraphics[width=0.9\columnwidth]{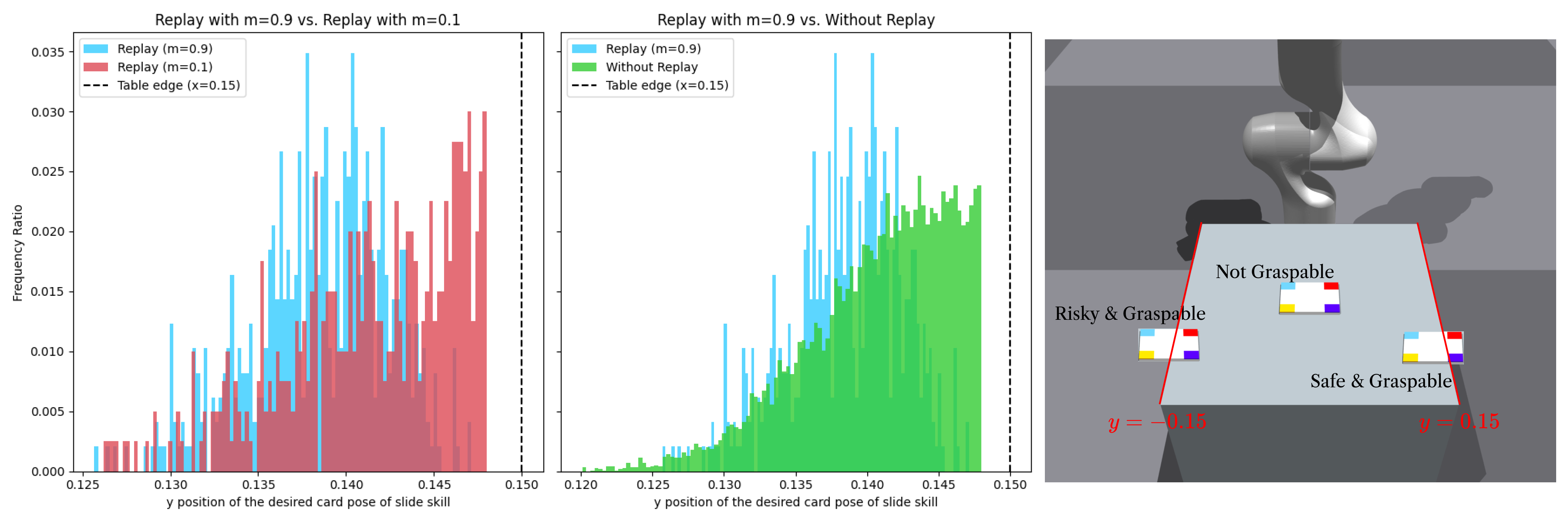}
\caption{The charts on the left and center compare the absolute value of the \(y\)-position of the desired card pose for the slide skill (before the P skill) across each filtering method. The chart on the right shows the graspability of the card pose based on the \(y\)-position in the simulation.} 
\label{fig:card_y_position_histograms} 
\end{figure}

\begin{figure}[h] 
\centering
\includegraphics[width=0.9\columnwidth]{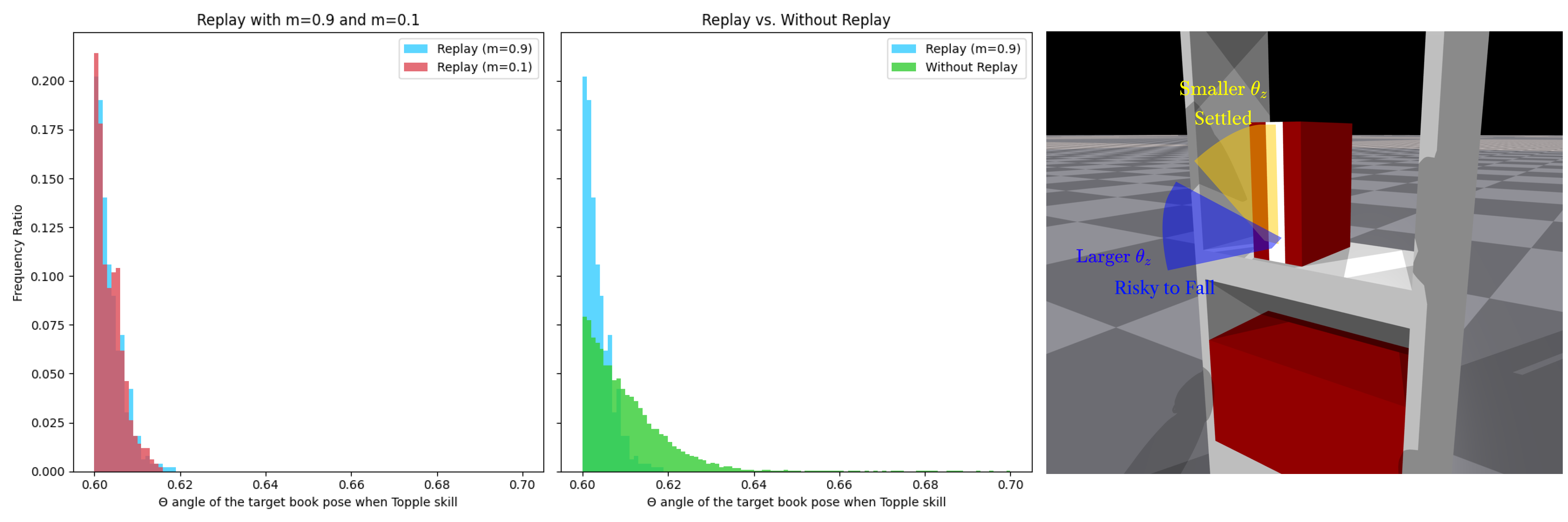}
\caption{The charts on the left and center compare the \( \theta_z \) (yaw) angle of the desired book pose for the Topple skill across each filtering method. The chart on the right shows that higher values of \( \theta_z \) are riskier for the book to fall down in the simulation.} 
\label{fig:book_thetay_position_histograms} 
\end{figure}

\begin{figure}[h] 
\centering
\includegraphics[width=0.9\columnwidth]{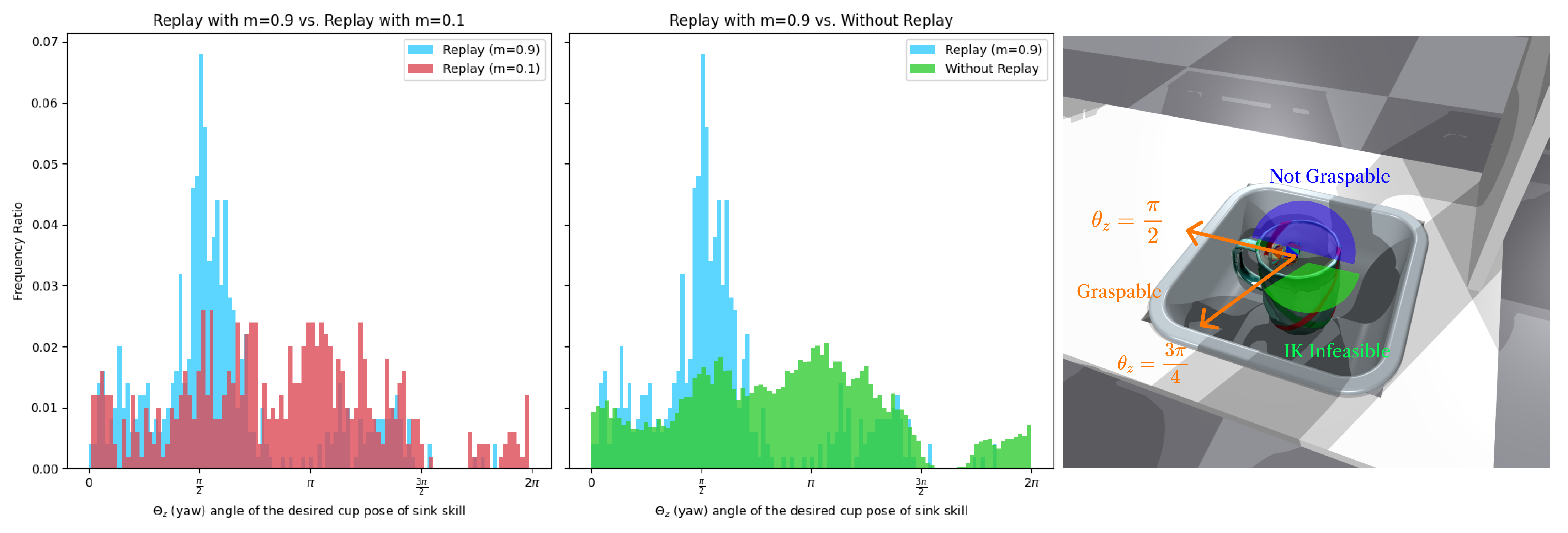}
\caption{The charts on the left and center compare the \( \theta_z \) (yaw) angle values of the desired cup pose for the Sink skill across each filtering method. The chart on the right shows the graspability based on the \( \theta_z \) (yaw) angle of the Sink skill's desired cup pose in the simulation.
} 
\label{fig:cup_theta_position_histograms} 
\end{figure}

We further analyze the collected skill plans from each method in two ways. First, we examine the distribution of replay success rates, as shown in Figure~\ref{fig:replay_success_dataQ}. Second, we investigate the desired object poses in the skill plans across the Card Flip, Bookshelf, and Kitchen domains, as presented in Figures~\ref{fig:card_y_position_histograms}, \ref{fig:book_thetay_position_histograms}, and \ref{fig:cup_theta_position_histograms}. Detailed explanations for each figure follow in the subsequent paragraphs.

Figure~\ref{fig:replay_success_dataQ} compares the replay success rates of skill plans collected by each filtering method. The y-axis shows the frequency of skill plans, and the x-axis represents their replay success rate. All success rates are measured by replaying each skill plan \( N = 400 \) times in the GPU-based simulator, IsaacGym. Filtering with \( m = 0.9 \) prioritizes high-quality skill plans by selecting those with a higher probability of successful replay. As a result, skill plans collected with \( m=0.9 \) exhibit a significantly higher average replay success rate compared to other methods.

The left and center charts of Figure~\ref{fig:card_y_position_histograms} show the distribution of the absolute \( y \)-position of the desired card pose for the slide skill, which is executed before the prehensile skill in the skill plan, extracted from skill plans collected by each filtering method. The y-axis shows the frequency, and the x-axis represents the absolute \( y \)-position. For successful grasping, the card should be placed near the table edge (\( y = 0.15 \) or \( -0.15 \)), but placing it too close increases the risk of falling. Thus, the ideal \( y \)-position is slightly smaller than 0.15, balancing graspability and stability. The right chart of Figure~\ref{fig:card_y_position_histograms} confirms that skill plans with higher success rates position the card further inward, reducing failure risk while maintaining graspability.

The left and center charts of Figure~\ref{fig:book_thetay_position_histograms} show the distribution of the $\theta_z$ (yaw) angle of the desired book pose for the topple skill. The y-axis indicates frequency, and the x-axis represents the $\theta_z$ angle. A larger toppling angle increases the risk of the book falling, leading to task failure, as shown in the right chart. Skill plans with higher success rates constrain the topple angle to smaller values, ensuring stability, whereas plans with lower success rates allow larger angles, increasing instability.

The left and center charts of Figure~\ref{fig:cup_theta_position_histograms} show the distribution of the $\theta_z$ (yaw) angle of the desired cup pose for the sink skill. The y-axis shows frequency, and the x-axis represents the $\theta_z$ angle. For efficient execution, the cup’s handle should ideally point toward the robot within \( [\pi/2, 3\pi/4] \). If the handle is not properly oriented, grasping becomes physically infeasible: when the handle faces away, the end-effector cannot reach the grasp point due to collisions with the underside of the cupboard. Conversely, even if the handle faces the robot directly, it may still be difficult to find a collision-free grasp pose because of the robot hardware constraints, resulting in inverse kinematics (IK) failure. The right chart shows that skill plans with higher success rates fall within the optimal range, whereas lower-quality plans cover a broader, suboptimal range.

While filtering with a high replay success threshold (\( m = 0.9 \)) leads to robust distillation policy performance, it also increases collection time, as shown in Table~\ref{table:ablation_dataQ_time}. Many skill plans are discarded for not meeting the \( m=0.9 \) criterion, resulting in longer collection time per plan. To mitigate this, we generate 30 successful trajectories per skill plan by replaying, leveraging the simulator's stochasticity and observation noise from domain randomization. This approach is more time-efficient than collecting a large number of skill plans, as done in the Without Replay setting.

\begin{table*}[ht]
\begin{adjustbox}{width=\columnwidth} 

\centering

\begin{tabular}{ccc|ccc|ccc|ccc}
    \toprule
    & \multicolumn{2}{c}{Data Collection} & \multicolumn{3}{c}{Card Flip} & \multicolumn{3}{c}{Bookshelf} & \multicolumn{3}{c}{Kitchen}   \\
    \cmidrule(lr){2-3} \cmidrule(lr){4-6}\cmidrule(lr){7-9}\cmidrule(lr){10-12}
    &  \makecell{Skill \\ plans} & \makecell{Traj \\ per a plan} & \makecell{Time \\ per plan (s)} & \makecell{Total \\ time (s)} & \makecell{Discarded \\ plans} &  \makecell{Time \\ per plan (s)} & \makecell{Total \\ time (s)} & \makecell{Discarded \\ plans}  & \makecell{Time \\ per plan (s)} & \makecell{Total \\ time (s)}& \makecell{Discarded \\ plans} \\
    \cmidrule(lr){1-1} \cmidrule(lr){2-3} \cmidrule(lr){4-6}\cmidrule(lr){7-9}\cmidrule(lr){10-12}
    \makecell{\textit{Without}\\ \textit{Replay}} 
    & 15000
    & 1
    & 85.3
    & 1.28 * $10^6$
    & 0
    & 79.2
    & 1.19 * $10^6$
    & 0
    & 121.0
    & 1.82 * $10^6$
    & 0
    \\
    \midrule
    \makecell{\textit{Replay} \\\textit{with $(m=0.1)$}} 
    & 500
    & 30
    & 89.5
    & 4.47 * $10^4$
    & 23
    & 79.5
    & 3.98 * $10^4$
    & 2
    & 122.6
    & 6.13 * $10^4$
    & 6
    \\
    \midrule
    \makecell{Ours (\textit{Replay}\\ \textit{with $(m=0.9)$)}} 
    & 500
    & 30
    & 2372.1
    & 1.18 * $10^6$
    & 13403
    & 449.4
    & 2.2 * $10^5$
    & 2437
    & 1048.6
    & 5.2 * $10^5$
    & 3933
    \\
    \bottomrule
\end{tabular}
\end{adjustbox}

\caption{Data collection time for each filtering method. "Skill plans" refers to the number of skill plans for each filtering method. ”Traj per a plan” refers to the number of successful trajectories collected for each skill plan. "Time per plan" indicates the time taken to collect a skill plan using the planner \texttt{Skill-RRT}. "Total time" represents the total collection time required to gather the specified number of skill plans under each filtering method. "Discarded plans" denotes the number of skill plans discarded due to failing to meet the replay success rate criteria (\( m = 0.1 \) or \( m = 0.9 \)).}
\label{table:ablation_dataQ_time}
\end{table*}

\label{Appendix:data_qual}

\subsection{Imitation Leaning Policy Architectures}
In this section, we study how the policy architecture choices in IL affect the performance of the distillation policy in simulation. Specifically, we compare the diffusion policy~\cite{chi2023diffusion} with four alternative architectures: (1) ResNet~\cite{he2016deep}, a simple IL model with a large parameter size; (2) LSTM+GMM, which has been shown to be effective for multimodal data in RoboMimic~\cite{robomimic2021}; (3) cVAE~\cite{kingma2013auto}, another conditional generative model; and (4) Transformer~\cite{vaswani2017attention}, a widely used architecture for multimodal data such as language. Each architecture is trained on the same dataset, with approximately 70 million parameters, for 100 epochs, across three different seeds.

For training, we adapt the IL codebase from RoboMimic~\cite{robomimic2021}. The MLP backbone is replaced with ResNet for a fair comparison with other large models. We do not modify the original code of LSTM+GMM, cVAE, or Transformer, except for adjusting hyperparameters such as model size. The hyperparameters, which are kept consistent across the three domains, are summarized in Table~\ref{table:IL_ablation_hyper}. Each architecture is trained for 100 epochs.

\begin{table}[h!]
    \centering

    \setlength\tabcolsep{9 pt}
    \begin{adjustbox}{width=0.8\textwidth} 
    \begin{tabular}{c|c|c|c}
    \toprule
    \multicolumn{4}{c}{Common} \\
    \midrule
    State Normalization & Z-Score & Action Normalization & Min-Max \\
    
    Batch size & 2048 & Epochs & 100 \\
    
    Optimizer & AdamW & Learning Rate & 1e-4 \\
    
    Weight Decay & 0.1 & LR Scheduler & Cosine \\
    
    \midrule
    \multicolumn{2}{c|}{ResNet} & \multicolumn{2}{c}{LSTM+GMM} \\
    
    \midrule
    MLP Dimensions & [4096x3, 2048x7, 1024]  & 
    History Length & 10 \\
    
    Activation Function & GELU & 
    LSTM Dimensions & [1470x4] \\
    
    Loss & L2 Norm & 
    Last Layer Dimension & 2048 \\
    
    & & 
    Activation Function & GELU \\
    
    & & 
    Number of GMM Modes & 10 \\
    
    & & 
    Loss & NLL \\
    \midrule
    \multicolumn{2}{c|}{cVAE} & \multicolumn{2}{c}{Transformer} \\
    \midrule
    Encoder Dimensions & [1024, 1024, 2048] & 
    Embedding Dimension & 1024 \\
    Decoder Dimensions & [1024, 1024, 2048] & 
    Number of Layers & 6 \\
    Prior Dimensions & [1024, 1024, 2048] & 
    Number of Heads & 8 \\
    Latent Dimensions & 14 & 
    Dropout & 0.1 \\
    Loss & VAE & 
    Activation & GELU \\
    KL Loss Weight & 1.0 & 
    Loss & L2 \\
    \bottomrule
    \end{tabular}
    \end{adjustbox} 
    \caption{Hyperparameters of IL policy architectures}
    \label{table:IL_ablation_hyper}
\end{table}
\begin{table}[h!]
    \centering
    \setlength\tabcolsep{12 pt}
    \begin{adjustbox}{width=0.6\textwidth} 
    \begin{tabular}{r|ccc}
    \toprule
                        & Card Flip & Bookshelf & Kitchen \\
    \midrule
    ResNet \cite{he2016deep}                    & \small 54$\, \pm \,$5.7  & \small 85$\, \pm \,$2.6  & \small 98$\, \pm \,$0.5 \\ 
    LSTM+GMM \cite{robomimic2021}               & \small 55$\, \pm \,$5.7  & \small 95$\, \pm \,$1.7  & \small \textbf{99$\, \pm \,$0.5} \\ 
    cVAE \cite{kingma2013auto}                  & \small 22$\, \pm \,$2.1  & \small 31$\, \pm \,$3.3  & \small 41$\, \pm \,$4.5 \\ 
    Tranformer    \cite{vaswani2017attention}   & \small 63$\, \pm \,$1.4  & \small \textbf{96$\, \pm \,$1.4}  & \small 98$\, \pm \,$0.8 \\ 
    Diffusion Policy  \cite{chi2023diffusion}   & \small \textbf{95$\, \pm \,$0.5}   & \small 93$\, \pm \,$1.4  & \small 98$\, \pm \,$0.5 \\ 
    \bottomrule
    \end{tabular}
    \end{adjustbox} 
    \caption{Success rates with unit (\%) averaged over 3 seeds in simulation for each IL policy architectures.}
    \label{table:IL_arch_ablation}
\end{table}

The success rate is measured on 100 fixed test problems, which specify the initial object configuration $s_0.q_{\text{obj}}$, the initial robot configuration $s_0.q_{\text{r}}$, and the goal object configuration $q^g_{\text{obj}}$. We record the success rate at every epoch, and the performance of each architecture is evaluated based on the maximum success rate achieved during training across three seeds. The results are summarized in Table~\ref{table:IL_arch_ablation}.

In the card flip domain, the diffusion policy significantly outperforms the other architectures. We hypothesize that this advantage stems from the high diversity of intermediate target poses in this domain, where the object poses can vary between, for example, the left and right edges of the table. In contrast, in the bookshelf and kitchen domains, ResNet, LSTM+GMM, and Transformer achieve success rates comparable to the diffusion policy, possibly due to the lower diversity of intermediate poses. cVAE exhibits consistently poor performance across all domains, suggesting that the smoothing effect of VAE impairs IL performance. These results indicate that the overall success of our framework is not solely attributed to the choice of the diffusion policy. Although the diffusion policy achieves slightly higher performance, other architectures such as ResNet, LSTM+GMM, and Transformer also demonstrate strong success rates. This suggests that our framework's effectiveness is not dependent on a specific policy architecture, but rather that the diffusion policy was selected as the optimal choice among several viable options.\label{Appendix:IL_arch_ablation}
\end{document}